%% file: arxiv-new.tex
\documentclass{article}

 \usepackage[eandd, final]{neurips_2026}

\usepackage[utf8]{inputenc} 
\usepackage[T1]{fontenc}    
\usepackage{hyperref}       
\usepackage{url}            
\usepackage{booktabs}       
\usepackage{amsfonts}       
\usepackage{nicefrac}       
\usepackage{microtype}      
\usepackage[table]{xcolor}  
\usepackage{graphicx}
\usepackage{multirow}
\usepackage{makecell}
\usepackage{diagbox}
\usepackage[export]{adjustbox}  
\usepackage{enumitem}  
\usepackage{amssymb}

\definecolor{scoreMaxColor}{RGB}{170,215,185}
\newcommand{\rgb}[1]{\cellcolor{scoreMaxColor!#1!white}#1}
\definecolor{artifactblue}{RGB}{63,145,201}

\title{\textsc{SalArt-VQA}: Diagnosing Whether VLMs Understand Salient Artifacts in Generated Images
}

\author{%
Ruotian Zhang\thanks{Equal contribution.}\\
Zhejiang University \\
\And
Xiaoxiao Sun$^{*}$  \\
Stanford University\\
  \AND
  Junzhe Huang \\
The University of Queensland\\
  \And
James Burgess \\
Stanford University\\
  \And
Serena Yeung-Levy\\
Stanford University \\
}

\begin{document}

\maketitle
\vspace{-1.5em} \begin{center} 
\raisebox{-0.2em}{\includegraphics[height=1.1em]{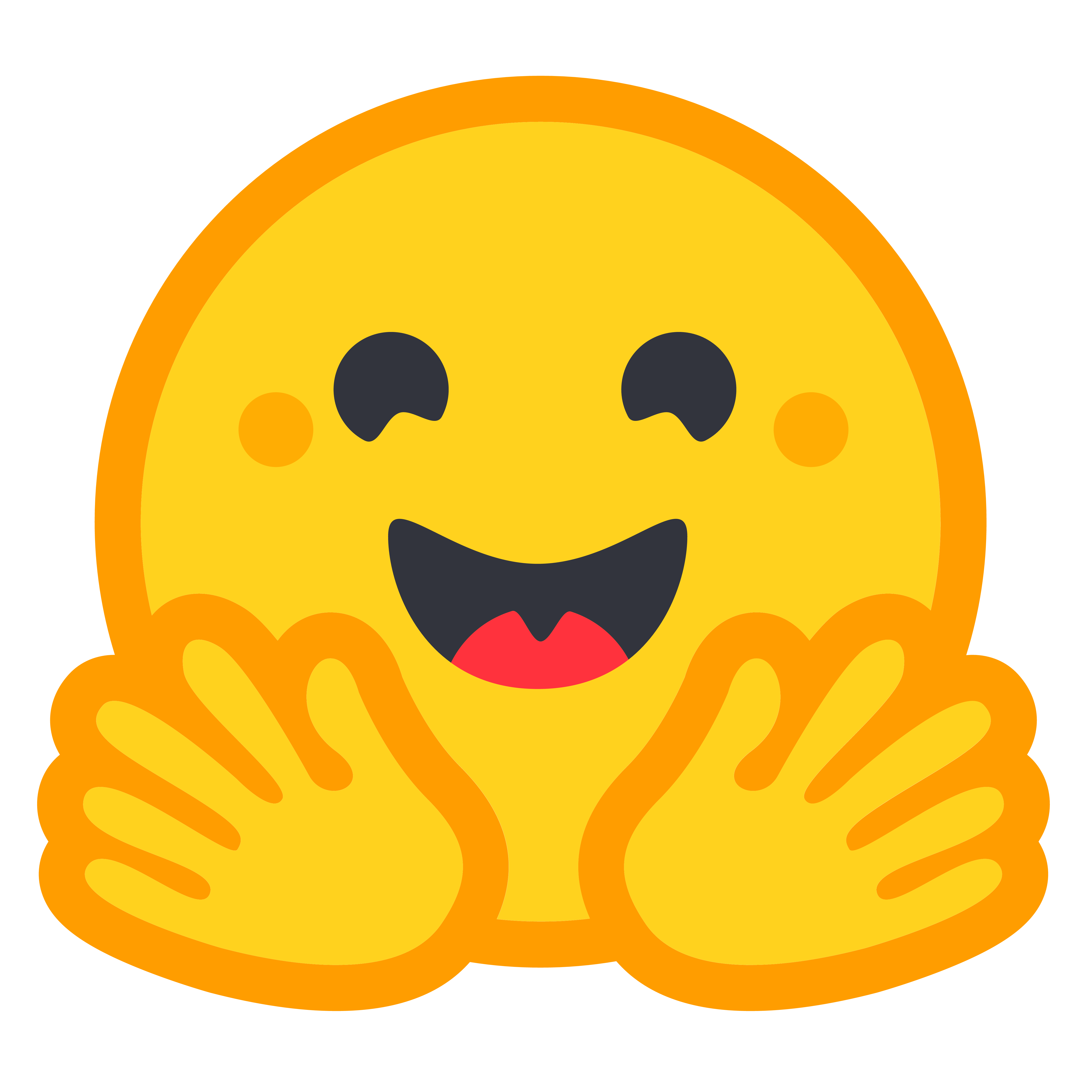}} \ \href{https://huggingface.co/datasets/salartvqa/SalArt-VQA} {\texttt{https://huggingface.co/datasets/salartvqa/SalArt-VQA}} 

\raisebox{-0.2em}{\includegraphics[height=1.1em]{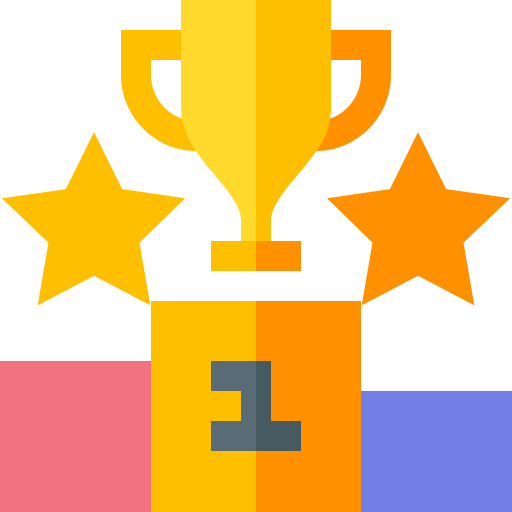}}
\ \href{https://zhangruotian.github.io/SalArt-VQA/}{\texttt{https://zhangruotian.github.io/SalArt-VQA/}}
\end{center} \vspace{0.5em}

\begin{abstract}
Vision-language models (VLMs) are increasingly used to detect whether AI-generated images contain visible artifacts, yet their ability to analyze such artifacts remains poorly understood. A correct image-level decision can still hide important failures: a model may correctly flag an artifact while relying on the wrong visual cue, selecting the wrong region, or describing a defect that the image does not support. To evaluate these behaviors directly, we introduce \textsc{SalArt-VQA}, a diagnostic benchmark for fine-grained \textbf{SAL}ient \textbf{ART}ifact understanding in AI-generated images. \textsc{SalArt-VQA} contains 950 images and 3,681 human-authored multiple-choice questions spanning artifact images, matched real reference images, and paired generated reference images. Four aligned question types evaluate presence detection, semantic localization, spatial grounding, and evidence-grounded defect identification, while the reference splits test calibration and abstention when the annotated defect is absent. Across 20 VLMs, \textsc{SalArt-VQA} reveals failures that image-level detection accuracy hides: the strongest model reaches 99.37\% detection recall on artifact images but answers all four artifact-side questions correctly on only 53.26\% of images. Comparing artifact images with artifact-free references reveals a sensitivity-calibration tradeoff: sensitive models often make unsupported artifact claims, while conservative models avoid false alarms largely by missing real artifacts. These results show that high artifact detection accuracy alone does not imply grounded artifact understanding. \textsc{SalArt-VQA} exposes these hidden failure modes and provides a fine-grained evaluation of whether VLM artifact claims are supported by local visual evidence. 

\end{abstract}

\section{Introduction} 
\label{sec:intro}
As AI-generated images improve and spread across real-world applications~\citep{muthaiah2024comparative, rombach2022high}, automatically determining whether an image contains visible artifacts is becoming increasingly important~\citep{baraheem2023ai}. Such artifacts may appear as structural distortions, inconsistent object geometry, broken reflections, implausible spatial relations, or malformed text. Although many salient artifacts are visually prominent to human viewers, automatically analyzing them remains challenging: a system must recognize subtle irregularities, describe open-ended defect types, and ground each claim to the supporting image region. Vision-language models (VLMs) are therefore increasingly used for artifact detection~\citep{burgess2026artifactlens, wang2024detecting}, and as evaluators in generative-image pipelines that screen outputs, rank candidate generations, or provide feedback for refinement. However, their actual ability to reliably perceive, localize, and explain visible artifacts remains poorly understood. This gap matters: unsupported artifact claims can reject valid images, misdirect edits, or mislead users.

\begin{figure}[t]
  \centering
  \includegraphics[width=\textwidth]{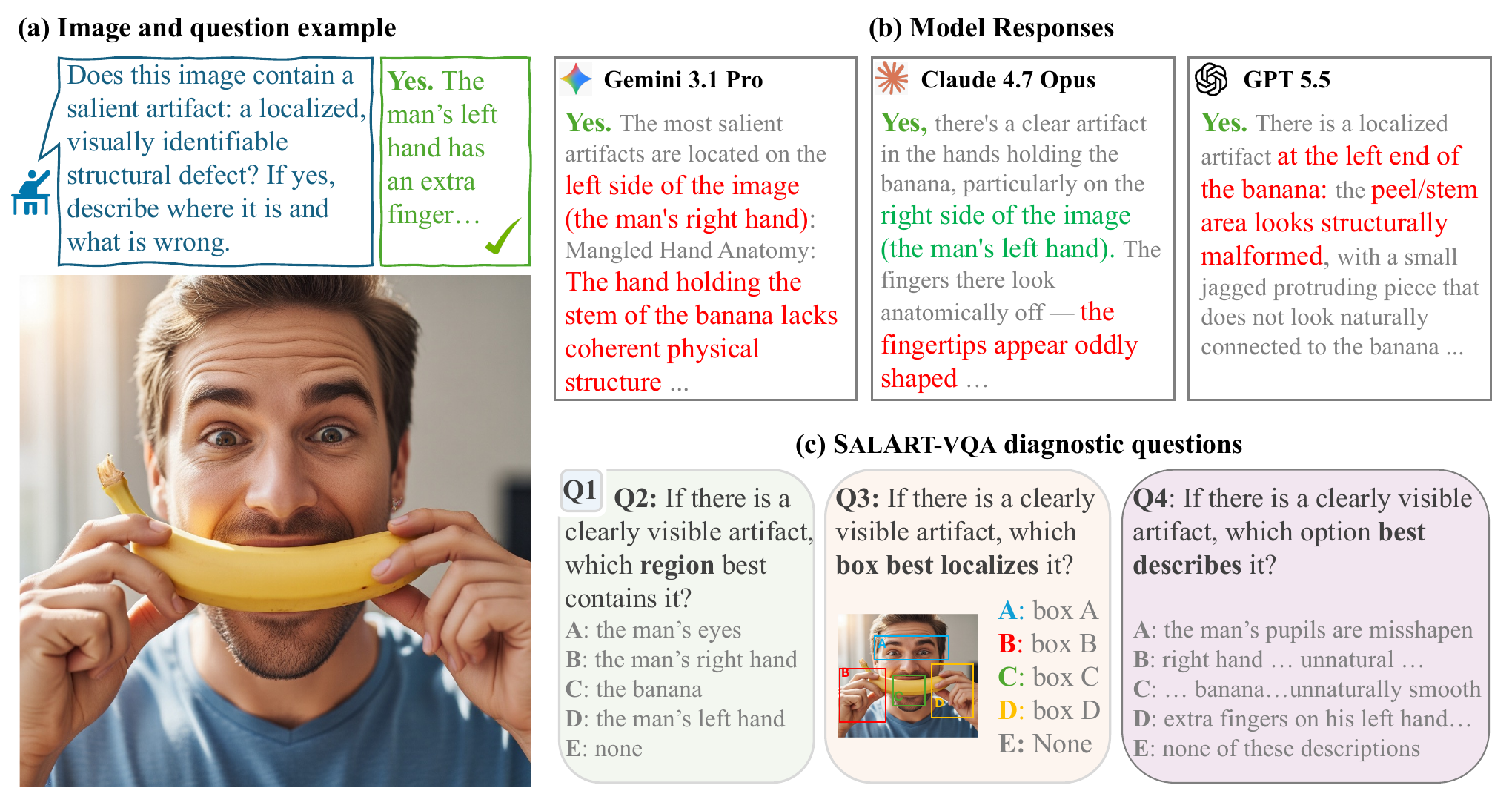}
\caption{\textbf{Motivation and question example for \textsc{SalArt-VQA}}. Image-level artifact detection may look successful even when downstream behavior differs substantially. In panels (a) and (b), several VLMs flag the image as artifacted, but most refer to the wrong region and none gives the correct defect description. Panel (c) shows how \textsc{SalArt-VQA} probes artifact presence (Q1), location (Q2-Q3), and visual evidence (Q4) beyond the artifact-present decision.}
  \label{fig:overview}
\end{figure}

As shown in Figure~\ref{fig:overview}(a,b), these risks are not captured by image-level artifact detection alone. In the example, several models correctly answer ``yes'' to the presence question, yet their follow-up responses point to different regions and different defect hypotheses. A model may therefore make the right image-level decision while relying on the wrong visual cue, such as selecting the wrong object part, confusing which local abnormality is responsible, or describing a defect that the image does not support. These failures matter because artifact analysis is often used to guide downstream actions, where the predicted location and explanation can be as important as the binary artifact-present decision. To study these behaviors more directly, we introduce \textsc{SalArt-VQA}, a diagnostic benchmark for salient artifact\footnote{We use \emph{salient artifact} to refer to a human-visible defect that is visually prominent, localizable, and describable in natural language; the benchmark focuses on such target defects.} understanding in AI-generated images.

\textsc{SalArt-VQA} is designed to expose these distinctions through aligned, closed-set VQA questions. The questions form a diagnostic chain: they ask whether a model detects artifact presence, names the relevant region, grounds that region spatially, and selects the defect description supported by the image (Figure~\ref{fig:overview}(c)). The benchmark contains 950 images and 3,681 human-annotated multiple-choice questions spanning artifact images, matched real references, and paired generated references. Artifact images evaluate model behavior on genuine salient defects; matched real references test calibration on real photographs; and paired generated references test whether models can abstain from artifact claims when the annotated salient defect is absent from a visually similar generated scene.

Using these diagnostics, we evaluate 20 VLMs and several specialized artifact-analysis baselines. On artifact images, current VLMs struggle with the full artifact-understanding chain: the strongest model answers all four questions correctly on only 53.3\% of images, and the next-best model remains near 33\%, despite much higher artifact-presence detection accuracy. Across artifact and reference images, we further find a clear sensitivity-calibration tradeoff: no model combines high artifact sensitivity with strong reference-side calibration. Current systems largely split into two main regimes: aggressive detectors that perform best on artifact images but make unsupported artifact claims on reference images, and conservative models that stay calibrated to reference images but are insensitive to artifacts. Additional analyses reveal finer-grained failures in localization, evidence selection, and reference-side abstention. \textsc{SalArt-VQA} makes these differences visible, exposing capability gaps that aggregate detection accuracy alone would hide. Our contributions are three-fold:
\begin{itemize}[leftmargin=1.2em, itemsep=0.2em, topsep=0.2em]
    \item We introduce \textsc{SalArt-VQA}, a closed-set VQA benchmark for fine-grained diagnosis of salient artifact understanding in AI-generated images beyond image-level detection.
    \item We design aligned question types together with matched real and paired generated reference images that separate artifact sensitivity, semantic localization, spatial grounding, evidence reasoning, and reference-side calibration.
    \item We evaluate 20 VLMs and several specialized methods, showing that current systems struggle with the full artifact-understanding chain and exhibit a consistent sensitivity-calibration tradeoff.
\end{itemize}

\section{Related work}
\label{sec:related}

\textbf{Synthetic image detection benchmarks.}
Much prior work treats AI-generated image detection as binary classification and studies generalization across generators or deployment conditions. GenImage~\citep{zhu2023genimage}, B-Free~\citep{guillaro2025bias}, AIDE~\citep{yan2024sanity}, and RRDataset~\citep{li2025bridging} evaluate synthetic-image detection under cross-generator, bias, fidelity, and real-world transmission shifts. LOKI~\citep{ye2024loki} introduces a benchmark for synthetic data detection, evaluating whether large multimodal models can distinguish real from AI-generated content across multiple modalities. In contrast, our work focuses on visible artifact understanding in AI-generated images, assessing whether VLMs can detect, localize, and explain concrete visual defects.

\textbf{Fine-grained and explainable artifact analysis.}
Other work moves toward artifact localization, categorization, and explanation. PAL~\citep{zhang2023perceptual} provides per-pixel artifact labels; HiFi-IFDL~\citep{guo2023hierarchical} and TruFor~\citep{guillaro2023trufor} localize manipulated regions; HAD~\citep{wang2024detecting} targets anatomical anomalies; and related studies build perceptual-artifact taxonomies or human-visible artifact categories~\citep{xiao2026unveiling,kamali2025characterizing}. LEGION~\citep{kang2025legion} combines detection, segmentation, and explanation; AIGI-Holmes~\citep{zhou2025aigi} and FakeXplain~\citep{jifakexplain} train localized rationales; and ArtifactLens~\citep{burgess2026artifactlens} adapts pretrained VLMs for low-label artifact recognition. Rather than training artifact recognizers or collecting free-form rationales, \textsc{SalArt-VQA} provides a closed-set diagnostic benchmark for off-the-shelf VLMs, linking detection, localization, grounding, evidence selection, and reference-side abstention for the same salient artifact.

\textbf{Grounded and diagnostic multimodal evaluation.}
Artifact understanding also relates to region-grounded perception and diagnostic VQA. GLIP~\citep{li2022grounded}, Grounding DINO~\citep{liu2024grounding}, GPT4RoI~\citep{zhang2024gpt4roi}, GLaMM~\citep{rasheed2024glamm}, and LMM-Det~\citep{li2025lmm} study open-vocabulary or region-level grounding. VQA v2~\citep{goyal2017making}, BLINK~\citep{fu2024blink}, and HallusionBench~\citep{guan2024hallusionbench} use controlled questions or paired examples to probe perception and reasoning, while POPE~\citep{li2023evaluating}, NOPE~\citep{lovenia2024negative}, and removed-object counterfactual evaluation~\citep{he2025evaluating} test object hallucination. \textsc{SalArt-VQA} shifts this diagnostic style from object presence to artifact evidence, using aligned questions and paired references to test whether a claimed defect remains visually supported.

\section{\textsc{SalArt-VQA} Construction}
\label{sec:dataset}

\begin{figure}[t]
  \centering
  \includegraphics[width=\textwidth]{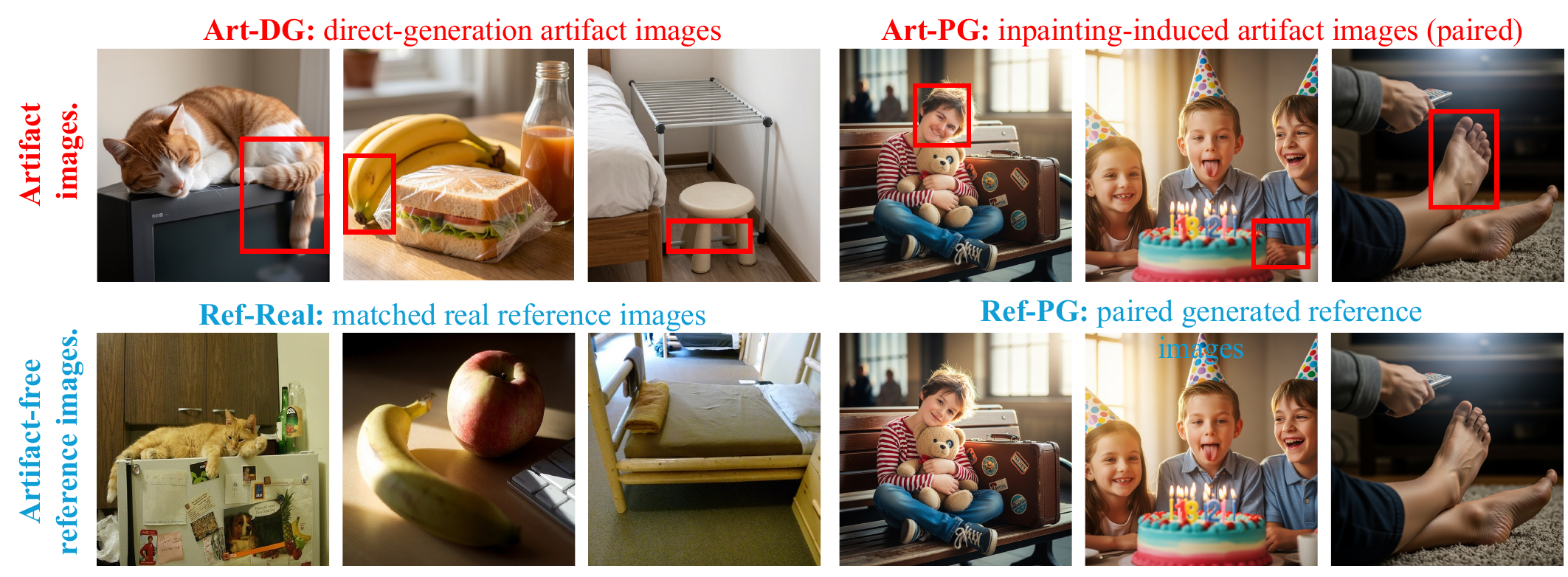}
\caption{\textbf{Examples from the \textsc{SalArt-VQA} image splits.} Artifact images from Art-DG and Art-PG are shown with their corresponding reference images from Ref-Real and Ref-PG.}
  \label{fig:examples}
\end{figure}

\subsection{Image Collection}
\label{sec:images-collection} 

\textsc{SalArt-VQA} contains four image splits: direct-generation artifact images (Art-DG), inpainting-induced artifact images from paired generated scenes (Art-PG), matched real reference images (Ref-Real), and paired generated reference images (Ref-PG). We use ``PG'' for the paired generated construction, where each Art-PG image is produced from a corresponding Ref-PG image.

\textbf{Artifact images.} These images evaluate whether a model can detect, localize, and correctly describe one annotated salient artifact. The benchmark questions target this selected artifact; in a small number of cases, weaker secondary imperfections may remain elsewhere. We keep only images with simple compositions where the selected artifact can be isolated by closed-set questions and described unambiguously in natural language. The artifact set combines two construction sources.

\hspace*{1em}\textit{(1) Direct-generation artifact images (Art-DG, 356).} We build this split primarily from images generated with Imagen~4~\citep{googledeepmind2025imagen4}, Midjourney~\citep{midjourney2026about}, and FLUX.2~[klein]~\citep{blackforestlabs2026flux2klein}. Prompts are adapted from COCO captions~\citep{lin2014microsoft} and augmented with photorealistic style cues such as ``natural light'' and ``detailed surroundings''; no prompt explicitly asks for artifacts. From around 13{,}000 images, we manually flag potential salient artifacts, remove non-photorealistic outputs, and keep only cases satisfying the criteria above.

\hspace*{1em}\textit{(2) Inpainting-induced artifact images (Art-PG, 119).}
These images form aligned Art-PG-Ref-PG pairs. Starting from a generated reference image, we edit one local body region while leaving the rest of the image unchanged. The pipeline uses MediaPipe~\citep{lugaresi2019mediapipe} to propose region hints, SAM~2~\citep{ravi2024sam} to refine them into masks, and FLUX.1-Fill-dev~\citep{blackforestlabs2026flux1filldev} to inpaint each mask with multiple random seeds. Annotators retain outputs with a single, clear salient artifact in the edited region. Appendix~\ref{app:injection} gives the full pipeline.

\textbf{Reference images.} These images support calibration checks in two complementary settings:

\hspace*{1em}\textit{(1) Matched real reference images (Ref-Real, 356).} These are ordinary real photographs that serve as real-image references for Art-DG. For each of the 356 Art-DG images, we retrieve a COCO photograph either using the same source caption or by selecting a high-similarity match in CLIP~\citep{radford2021learning} image-embedding space. We choose matches with similar visual content, including subject type, pose, and scene layout.

\hspace*{1em}\textit{(2) Paired generated reference images (Ref-PG, 119).}
These are the original generated images related to the Art-PG images, prior to the localized edit that introduces the target artifact. We do not assume that Ref-PG images are globally artifact-free; rather, they are artifact-free of the annotated salient defect targeted by the paired options. The correct answer for Q2-Q4 is ``none of these,''. This split supports paired comparisons between artifact-bearing images and generated references.

\textbf{Artifact taxonomy and subject categories.}
\label{sec:taxonomy}
After curation, we group the observed salient defects into five broad categories, informed by prior studies of perceptual artifacts~\citep{borji2023qualitative}: \emph{count anomaly} for wrong numbers of repeated parts such as fingers or teeth; \emph{topology anomaly} for merged, disconnected, intersecting, or clipped parts; \emph{local render/structure anomaly} for a single warped, smeared, or collapsed part; \emph{anatomy anomaly} for implausible large-scale body composition; and \emph{plausibility anomaly} for scene-level violations of physics, reflection, identity, or text. Images without a stable assignment are \emph{Others}. Figure~\ref{fig:statistics_questions}(a) shows the distribution and examples. These categories are descriptive metadata rather than answer labels.
Full definitions are given in Appendix~\ref{app:taxonomy}. The artifact images also span diverse subject categories, as shown by the common subject terms in Figure~\ref{fig:statistics_questions}(b).

\subsection{Multiple-Choice VQA Design}  
\label{sec:vqa}

\begin{figure}[t]
  \centering
  \includegraphics[width=\textwidth]{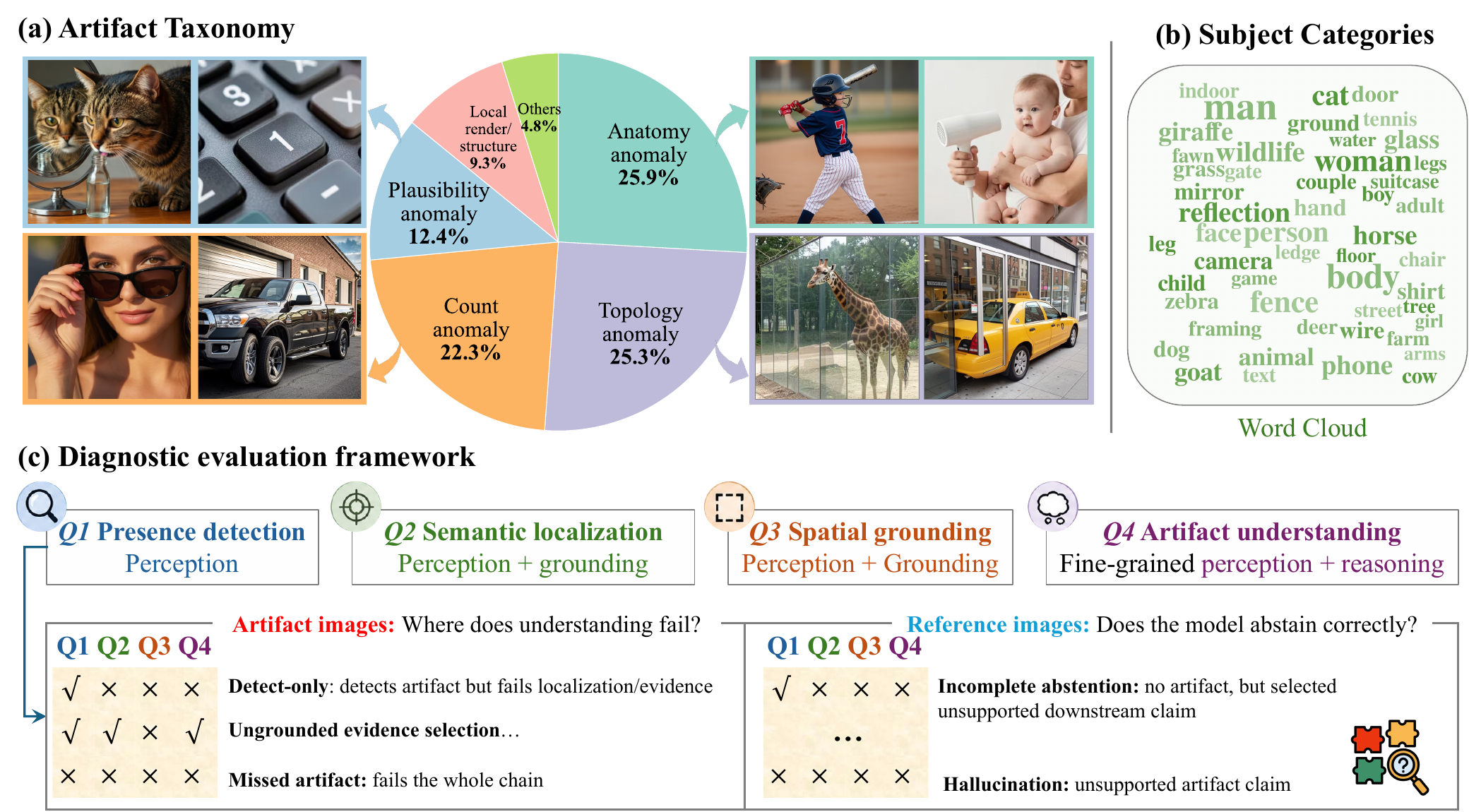}
\caption{\textbf{Dataset composition and diagnostic framework of \textsc{SalArt-VQA}.} (a) Curated artifact taxonomy with example images. (b) Common subject terms in the benchmark. 
(c) Diagnostic framework. The bottom panel shows examples of failure patterns. 
}
  \label{fig:statistics_questions}
\end{figure}

\textsc{SalArt-VQA} uses closed-set multiple-choice questions to reduce scoring ambiguity while keeping each answer tied to visible evidence. Figure~\ref{fig:overview}(c) and Figure~\ref{fig:statistics_questions}(c) summarize the four question types: Q1 asks whether a salient artifact is present, Q2 asks for the semantic region containing the artifact, Q3 asks for the corresponding bounding box, and Q4 asks for the defect description supported by the image. Full prompt templates and option conventions are given in Appendix~\ref{app:vqa_design}.

Options are constructed around the annotated target artifact rather than around taxonomy names. During annotation, candidate regions and defect hypotheses are initialized from image captions and auxiliary screening outputs, then manually verified. Annotators identify the target artifact region, insert or revise the correct option when needed, and retain plausible distractors only when they do not contain the target artifact and do not introduce ambiguous alternative answers. For Q2-Q4, the final A-D options follow an aligned design: the same letter refers to the same candidate region across semantic location, box grounding, and evidence description; option~E is always ``none of these.''

\textbf{Reference-image evaluation.}
The 356 Ref-Real images receive all four questions, with Q1\,=\,``no'' and Q2-Q4\,=\,E. The 119 Ref-PG images receive only Q2-Q4, also with answer~E, and \textbf{skip Q1}. Because Ref-PG images are themselves generated images, a global Q1 artifact-present judgment could be affected by generation imperfections unrelated to the paired salient defect. We therefore exclude Ref-PG from Q1 and use it only for the more targeted Q2-Q4 checks.

\textbf{Scoring and Diagnostic Evaluation.} Each question is presented in an independent, single-turn API call and scored by deterministic exact match. Beyond per-question accuracy, we analyze answer patterns across Q1-Q4, as shown in Figure~\ref{fig:statistics_questions}(c). For example, a model that answers Q1 and Q4 correctly but misses Q2 or Q3 selects the correct evidence statement but fails one of the explicit localization checks;  models that select A-D on references make unsupported artifact claims. These help separate detection,  localization,  grounding, evidence selection, and reference-side abstention.

\begin{table}[t]
  \caption{\textbf{Main results on \textsc{SalArt-VQA} (\%).} Columns are grouped by artifact images, reference images, and overall performance. Models are grouped by access category and sorted within each group by artifact-image All$_{Q1-4}$ accuracy; best per column in \textbf{bold}.}
  \label{tab:main}
  \centering
  \scriptsize
  \setlength{\tabcolsep}{2pt}
\begin{tabular}{@{}l|rrr|rrr|r|rrr}
  \toprule
  \multirow{3}{*}{\diagbox[width=3.65cm,height=1.75cm,innerleftsep=2pt,innerrightsep=2pt]{\textbf{Model}}{\textbf{Question}}{\textbf{Data type}}}
  & \multicolumn{3}{c|}{\textbf{\textcolor{red}{Artifact images}}}
  & \multicolumn{4}{c|}{\textbf{\textcolor{artifactblue}{Reference images}}}
  & \multicolumn{3}{c}{\textbf{Overall}} \\
  \cmidrule(lr){2-4} \cmidrule(lr){5-8} \cmidrule(lr){9-11}
  & \multicolumn{3}{c|}{Artifact}
  & \multicolumn{3}{c|}{Ref-Real}
  & \multicolumn{1}{c|}{Ref-PG}
  & \multicolumn{3}{c}{All splits}  \\
  \cmidrule(lr){2-4} \cmidrule(lr){5-7} \cmidrule(lr){8-8} \cmidrule(lr){9-11}
  &
  \makecell[c]{{\tiny Q1}\\{\tiny detection}\\{\tiny recall}} &
  \makecell[c]{{\tiny All$_{Q1-4}$}\\{\tiny complete}\\{\tiny accuracy}} &
  \makecell[c]{{\tiny All$_{Q2-4}$}\\{\tiny grounding}\\{\tiny accuracy}} &
  \makecell[c]{{\tiny Q1}\\{\tiny no artifact}\\{\tiny specificity}} &
  \makecell[c]{{\tiny All$_{Q1-4}$}\\{\tiny complete}\\{\tiny accuracy}} &
  \makecell[c]{{\tiny All$_{Q2-4}$}\\{\tiny grounding}\\{\tiny accuracy}} &
  \makecell[c]{{\tiny All$_{Q2-4}$}\\{\tiny grounding}\\{\tiny accuracy}} &
  \makecell[c]{{\tiny Q1}\\{\tiny detection}\\{\tiny accuracy}} &
  \makecell[c]{{\tiny All$_{Q1-4}$}\\{\tiny complete}\\{\tiny accuracy}} &
  \makecell[c]{{\tiny All$_{Q2-4}$}\\{\tiny grounding}\\{\tiny accuracy}} \\
  \midrule
  \textit{Random} & \rgb{50.00} & \rgb{0.40} & \rgb{0.80} & \rgb{50.00} & \rgb{0.40} & \rgb{0.80} & \rgb{0.80} & \rgb{50.00} & \rgb{0.40} & \rgb{0.80} \\
\textit{Human} & \rgb{100.00} & \rgb{100.00} & \rgb{100.00} & \rgb{99.53} & \rgb{99.53} & \rgb{99.53} & \rgb{99.16} & \rgb{99.80} & \rgb{99.80} & \rgb{99.72} \\
  \midrule
  \multicolumn{11}{@{}l}{\textbf{Closed-source}} \\
  \midrule
  \includegraphics[width=0.3cm,valign=c]{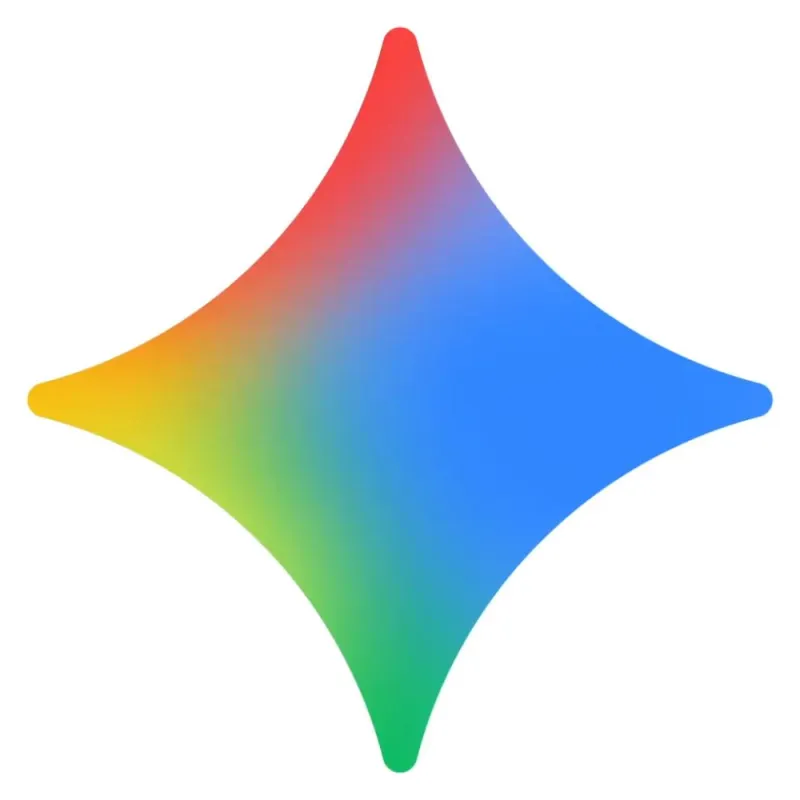}~Gemini 3.1 Pro          & \textbf{\rgb{99.37}} & \textbf{\rgb{53.26}} & \textbf{\rgb{53.89}} & \rgb{50.28} & \rgb{7.87}  & \rgb{8.71}  & \rgb{0.00} & \rgb{78.34} & \rgb{33.81} & \rgb{30.21} \\
  \includegraphics[width=0.3cm,valign=c]{fig-new/icons/gemini.png}~Gemini 3.1 Flash Lite   & \rgb{83.37} & \rgb{32.63} & \rgb{37.47} & \rgb{93.26} & \rgb{16.01} & \rgb{16.01} & \rgb{0.00} & \textbf{\rgb{87.61}} & \rgb{25.51} & \rgb{24.73} \\
  \includegraphics[width=0.3cm,valign=c]{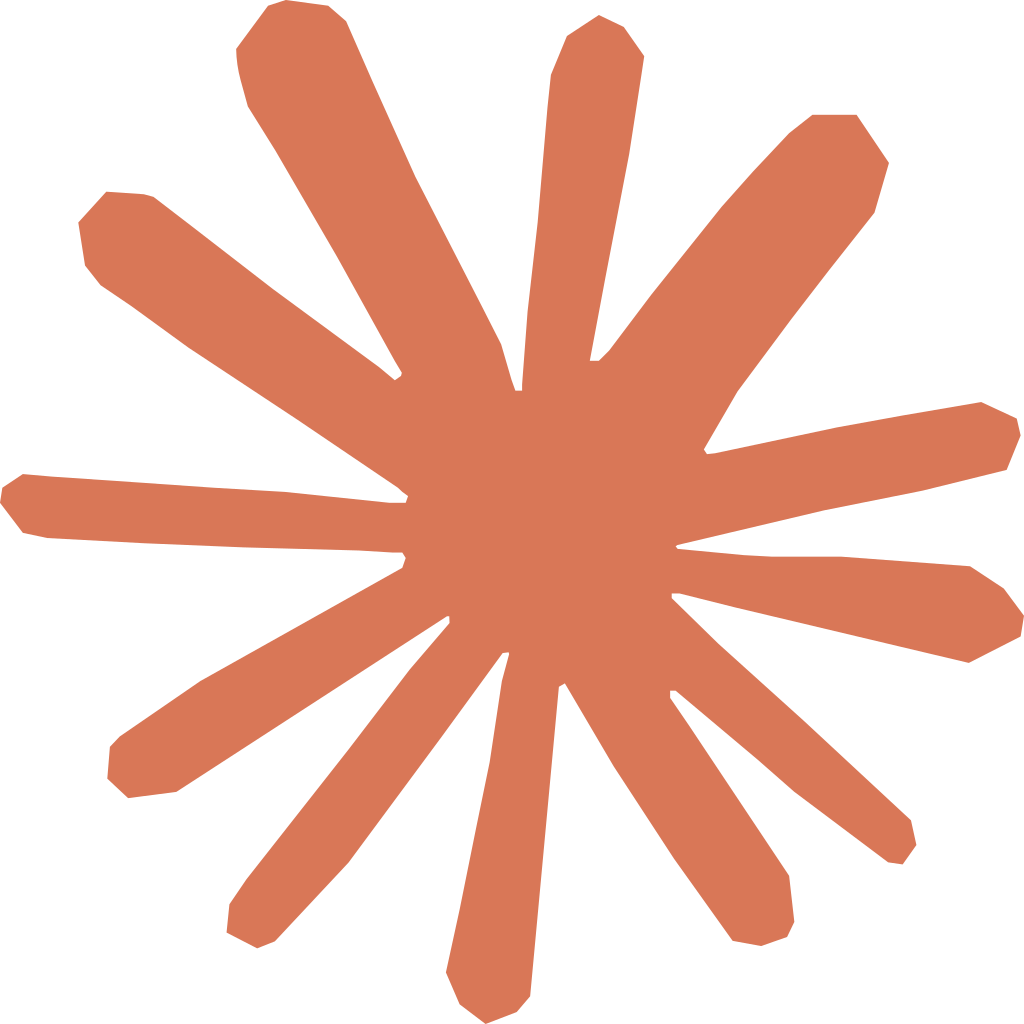}~Claude 4.6 Opus         & \rgb{47.58} & \rgb{11.79} & \rgb{17.47} & \textbf{\rgb{100.00}} & \rgb{96.35} & \rgb{96.35} & \rgb{0.84} & \rgb{70.04} & \textbf{\rgb{48.01}} & \rgb{44.95} \\
  \includegraphics[width=0.3cm,valign=c]{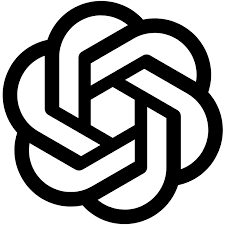}~GPT-5.4                 & \rgb{4.63}  & \rgb{1.89}  & \rgb{6.32}  & \textbf{\rgb{100.00}} & \rgb{95.79} & \rgb{95.79} & \rgb{20.17} & \rgb{45.49} & \rgb{42.12} & \rgb{41.58} \\
  \includegraphics[width=0.3cm,valign=c]{fig-new/icons/claude.png}~Claude 4.5 Haiku        & \rgb{0.42}  & \rgb{0.00}  & \rgb{0.84}  & \rgb{99.44} & \rgb{49.16} & \rgb{49.16} & \rgb{7.56} & \rgb{42.84} & \rgb{21.06} & \rgb{19.79} \\
  \includegraphics[width=0.3cm,valign=c]{fig-new/icons/openai.png}~GPT-5.4-nano            & \rgb{0.21}  & \rgb{0.00}  & \rgb{0.42}  & \rgb{99.72} & \rgb{78.65} & \rgb{78.93} & \rgb{19.33} & \rgb{42.84} & \rgb{33.69} & \rgb{32.21} \\
  \midrule
  \multicolumn{11}{@{}l}{\textbf{Open-weight (Thinking / Hybrid-thinking by default)}} \\
  \midrule
  \includegraphics[width=0.3cm,valign=c]{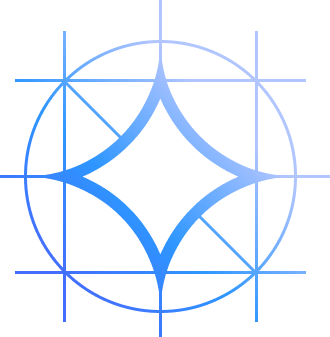}~Gemma-4-31B-it           & \rgb{86.32} & \rgb{33.26} & \rgb{37.68} & \rgb{68.26} & \rgb{3.93}  & \rgb{3.93}  & \rgb{0.00} & \rgb{78.58} & \rgb{20.70} & \rgb{20.31} \\
  \includegraphics[width=0.3cm,valign=c]{fig-new/icons/gemma.png}~Gemma-4-26B-it-A4B       & \rgb{57.89} & \rgb{17.68} & \rgb{28.00} & \rgb{83.71} & \rgb{14.61} & \rgb{15.17} & \rgb{0.00} & \rgb{68.95} & \rgb{16.37} & \rgb{19.68} \\
  \includegraphics[width=0.3cm,valign=c]{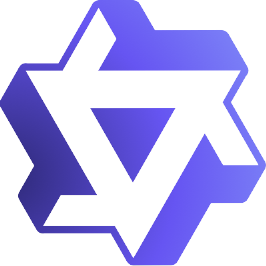}~Qwen3.5-397B-A17B         & \rgb{33.26} & \rgb{12.63} & \rgb{27.79} & \rgb{96.63} & \rgb{39.33} & \rgb{40.17} & \rgb{0.00} & \rgb{60.41} & \rgb{24.07} & \rgb{28.95} \\
  \includegraphics[width=0.3cm,valign=c]{fig-new/icons/qwen.png}~Qwen3.5-35B-A3B           & \rgb{20.63} & \rgb{6.32}  & \rgb{18.11} & \rgb{96.35} & \rgb{39.61} & \rgb{40.17} & \rgb{0.84} & \rgb{53.07} & \rgb{20.58} & \rgb{24.21} \\
  \includegraphics[width=0.3cm,valign=c]{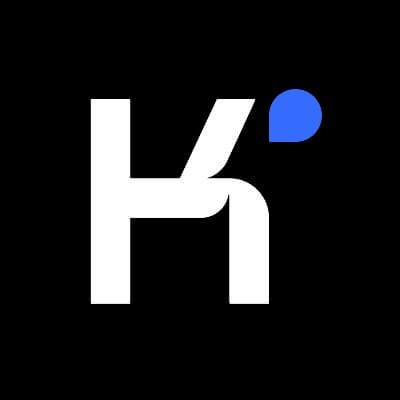}~Kimi K2.5                 & \rgb{3.79}  & \rgb{1.68}  & \rgb{4.84}  & \textbf{\rgb{100.00}} & \rgb{77.25} & \rgb{77.25} & \rgb{28.57} & \rgb{45.01} & \rgb{34.06} & \rgb{34.95} \\
  \includegraphics[width=0.3cm,valign=c]{fig-new/icons/qwen.png}~Qwen3-VL-30B-A3B-T        & \rgb{22.53} & \rgb{1.47}  & \rgb{3.16}  & \rgb{89.89} & \rgb{56.74} & \rgb{59.55} & \rgb{10.92} & \rgb{51.39} & \rgb{25.15} & \rgb{25.26} \\
  \includegraphics[width=0.3cm,valign=c]{fig-new/icons/qwen.png}~Qwen3-VL-235B-A22B-T      & \rgb{7.79}  & \rgb{1.26}  & \rgb{5.05}  & \rgb{96.91} & \rgb{35.67} & \rgb{35.96} & \rgb{11.76} & \rgb{45.97} & \rgb{16.00} & \rgb{17.47} \\
  \includegraphics[width=0.3cm,valign=c]{fig-new/icons/qwen.png}~Qwen3-VL-8B-T             & \rgb{3.79}  & \rgb{0.42}  & \rgb{2.74}  & \rgb{98.60} & \rgb{38.48} & \rgb{38.76} & \rgb{10.08} & \rgb{44.41} & \rgb{16.73} & \rgb{17.16} \\
  \includegraphics[width=0.3cm,valign=c]{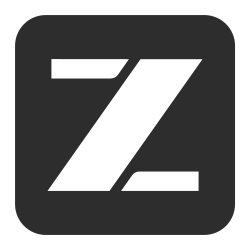}~GLM-4.6V                   & \rgb{0.42}  & \rgb{0.21}  & \rgb{0.21}  & \textbf{\rgb{100.00}} & \textbf{\rgb{97.47}} & \textbf{\rgb{97.47}} & \textbf{\rgb{84.87}} & \rgb{43.08} & \rgb{41.88} & \textbf{\rgb{47.26}} \\
  \midrule
  \multicolumn{11}{@{}l}{\textbf{Open-weight (Instruct)}} \\
  \midrule
  \includegraphics[width=0.3cm,valign=c]{fig-new/icons/qwen.png}~Qwen3-VL-30B-A3B-I        & \rgb{18.32} & \rgb{2.74}  & \rgb{7.16}  & \rgb{97.47} & \rgb{0.00}  & \rgb{0.00}  & \rgb{0.00} & \rgb{52.23} & \rgb{1.56} & \rgb{3.58} \\
  \includegraphics[width=0.3cm,valign=c]{fig-new/icons/qwen.png}~Qwen3-VL-235B-A22B-I      & \rgb{1.47}  & \rgb{0.42}  & \rgb{1.47}  & \textbf{\rgb{100.00}} & \rgb{92.70} & \rgb{92.70} & \rgb{84.03} & \rgb{43.68} & \rgb{39.95} & \rgb{46.00} \\
  \includegraphics[width=0.3cm,valign=c]{fig-new/icons/qwen.png}~Qwen3-VL-8B-I             & \rgb{0.42}  & \rgb{0.21}  & \rgb{1.05}  & \rgb{99.44} & \rgb{91.85} & \rgb{92.42} & \rgb{68.91} & \rgb{42.84} & \rgb{39.47} & \rgb{43.79} \\
  \makecell[l]{\includegraphics[width=0.3cm,valign=c]{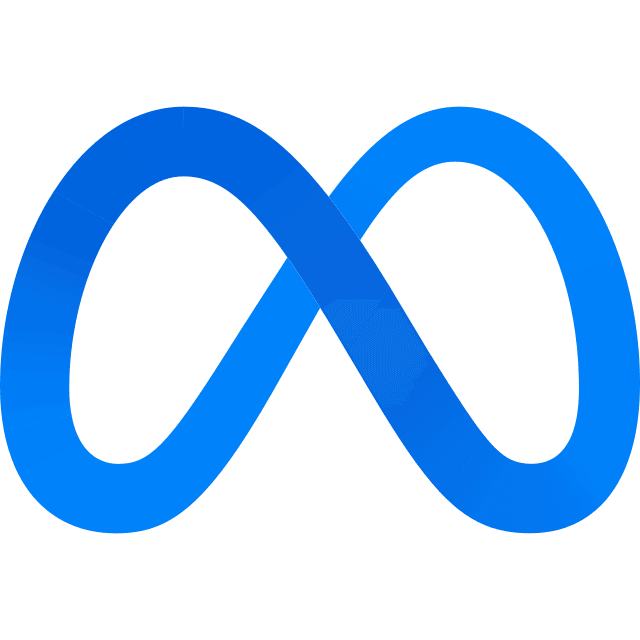}~Llama-4-Maverick-17B-128E-I} & \rgb{3.16} & \rgb{0.00} & \rgb{0.21} & \rgb{99.72} & \rgb{75.84} & \rgb{76.12} & \rgb{57.98} & \rgb{44.53} & \rgb{32.49} & \rgb{35.89} \\
  \makecell[l]{\includegraphics[width=0.3cm,valign=c]{fig-new/icons/Llama.png}~Llama-4-Scout-17B-16E-I}    & \rgb{0.63} & \rgb{0.00} & \rgb{0.00} & \textbf{\rgb{100.00}} & \rgb{82.30} & \rgb{82.30} & \rgb{74.79} & \rgb{43.20} & \rgb{35.26} & \rgb{40.21} \\
  \bottomrule
\end{tabular}
\end{table}

\section{Experiments}
\label{sec:experiments}

We evaluate 20 vision-language models from 9 families, covering both closed-source and open-weight systems. For clarity we group them by access category.

\begin{itemize}[leftmargin=1.2em, itemsep=0.2em, topsep=0.2em]
    \item \textbf{Closed-source:}
    \includegraphics[width=0.3cm,valign=c]{fig-new/icons/gemini.png} Gemini 3.1 Pro and Gemini 3.1 Flash Lite~\citep{gemini31pro};
    \includegraphics[width=0.3cm,valign=c]{fig-new/icons/claude.png} Claude Opus 4.6~\citep{claude_opus46} and Haiku 4.5~\citep{claude_haiku45};
    \includegraphics[width=0.3cm,valign=c]{fig-new/icons/openai.png} GPT-5.4 and GPT-5.4-nano~\citep{openai_gpt54}.

    \item \textbf{Open-weight (Thinking / Hybrid-thinking by default):}
    \includegraphics[width=0.3cm,valign=c]{fig-new/icons/qwen.png} Qwen3.5-397B-A17B and Qwen3.5-35B-A3B~\citep{qwen3.5};
    \includegraphics[width=0.3cm,valign=c]{fig-new/icons/gemma.png} Gemma-4-31B-it and -4-26B-it-A4B~\citep{gemma4modelcard};
    \includegraphics[width=0.3cm,valign=c]{fig-new/icons/zai.png} GLM-4.6V~\citep{vteam2025glm45vglm41vthinkingversatilemultimodal};
    \includegraphics[width=0.3cm,valign=c]{fig-new/icons/kimi.png} Kimi K2.5~\citep{kimiteam2026kimik25visualagentic};
    and \includegraphics[width=0.3cm,valign=c]{fig-new/icons/qwen.png}~Qwen3-VL-30B-A3B-Thinking, -235B-A22B-Thinking, and -8B-Thinking~\citep{qwen3technicalreport}.

    \item \textbf{Open-weight (Instruct):}
    \includegraphics[width=0.3cm,valign=c]{fig-new/icons/qwen.png} Qwen3-VL-30B-A3B-I, Qwen3-VL-235B-A22B-Instruct, and Qwen3-VL-8B-Instruct~\citep{qwen3technicalreport};
    \includegraphics[width=0.3cm,valign=c]{fig-new/icons/Llama.png} Llama-4-Maverick-17B-128E-Instruct and Llama-4-Scout-17B-16E-Instruct~\citep{llama4}.
\end{itemize}

In the following, we use \textbf{-I} and \textbf{-T} to denote Instruct-mode and Thinking-mode evaluations.

\textbf{Evaluation Protocol.} We evaluate all systems in a zero-shot, single-turn setting. Each MCQ is submitted independently with the corresponding image, fixed prompt wording, and image-specific options; no conversation history is shared across questions. For models that expose a temperature parameter, we set temperature to 0; otherwise, we use the provider default. Reasoning-capable models are queried under the default provider configuration unless an explicit Instruct/Thinking variant is named, and we do not manually override reasoning budgets. Llama models are evaluated using their publicly released multimodal Instruct weights. For the Human reference, three raters answer the same MCQs under the same single-turn, no-context protocol; we report mean accuracy across raters.

\subsection{Main Evaluation}
\label{sec:overview}

\textbf{Setup.} Table~\ref{tab:main} reports results on artifact images, matched real references (Ref-Real), and paired generated references (Ref-PG). Artifact and Ref-Real images receive Q1-Q4, while Ref-PG receives Q2-Q4 only, as described in Section~\ref{sec:vqa}. We report Q1 accuracy as artifact recall on artifact images and no-artifact specificity on Ref-Real, together with two image-level summaries: All$_{Q1-4}$ requires all four answers to be correct on the same image, and All$_{Q2-4}$ requires the three downstream answers to be correct. Rows are grouped by model access category and sorted within each group by artifact-image All$_{Q1-4}$. We first focus on the artifact-image block, which combines Art-DG and Art-PG (Appendix Table~\ref{tab:path_split} reports Art-DG and Art-PG separately), and then compare artifact and reference blocks to analyze behavior across data types.

\subsubsection{Artifact-Side Performance and Failure Patterns}
\label{sec:artifact-diagnostic}

\begin{figure}[t]
  \centering
  \includegraphics[width=\textwidth]{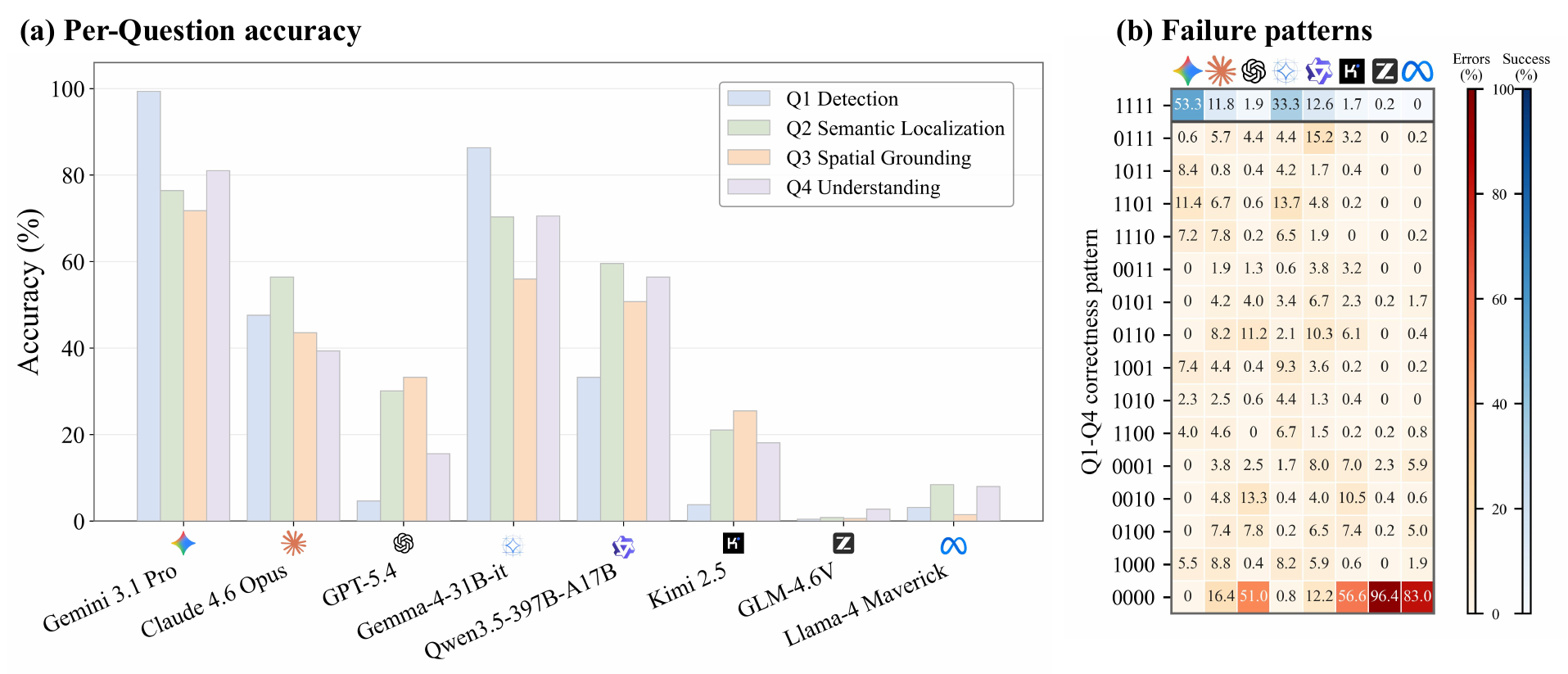}
  \caption{\textbf{Artifact-side diagnostic breakdown on representative models.} (a) Per-question accuracy on artifact images for Q1-Q4. (b) Image-level failure-pattern distribution over Q1-Q4 for the same models in the same left-to-right order, indicated by model icons. Here, \texttt{1} denotes a correct answer and \texttt{0} denotes an incorrect answer; \texttt{1111} indicates full success and \texttt{0000} indicates complete failure.}
  \label{fig:artifact_diagnostic}
\end{figure}

We first analyze the \textbf{artifact-image block} of Table~\ref{tab:main} and Figure~\ref{fig:artifact_diagnostic}.

\textbf{Even the binary presence decision is difficult for most models}. Only four VLMs exceed the 50\% random baseline on Q1 recall, namely Gemini~3.1~Pro, Gemini~3.1~Flash~Lite, Gemma-4-31B-it, and Gemma-4-26B-it-A4B. The remaining 16 models fall below chance on artifact detection, including several models that are strong on reference images. This shows that many current VLMs are conservative or insensitive to the salient artifacts.

\textbf{High artifact recall also does not translate into full artifact understanding.} Gemini~3.1~Pro reaches 99.4\% Q1 recall but answers all four artifact-side questions correctly on only 53.3\% of images. The next tier remains much lower: Gemma-4-31B-it and Gemini~3.1~Flash~Lite reach 33.3\% and 32.6\% All$_{Q1-4}$, respectively, despite strong Q1 recall. The same gap appears in the downstream-only metric: the best All$_{Q2-4}$ score is 53.9\%, followed by 37.7\%. Thus, detecting that an artifact is present is substantially easier than consistently identifying the relevant region, grounding it spatially, and selecting the supported defect description.

\textbf{Models with similar artifact-side performance tend to share similar bottlenecks.} Figure~\ref{fig:artifact_diagnostic}(a) shows this grouping at the per-question level. Gemini~Pro and Gemma-4-31B-it are strongest on Q1 and Q4, while their Q2 and Q3 accuracies are lower; for these models, detecting an artifact and selecting the supported defect description are easier than explicitly localizing the artifact by semantic region or bounding box. GPT-5.4, Kimi~K2.5, and Qwen3.5-397B-A17B show a different profile: their Q1 recall is much lower than several downstream accuracies, suggesting that they can sometimes identify the correct region or evidence under targeted Q2-Q4 options, even though they often do not commit to artifact presence. Claude~Opus~4.6 is distinct again, with Q2 semantic localization higher than its other artifact-side scores.

\textbf{Image-level answer patterns reveal where each model fails.} Figure~\ref{fig:artifact_diagnostic}(b) records which subset of Q1-Q4 is correct on the same image. Gemini~Pro has no \texttt{0000} (complete failure) cases; most of its errors are partial, and many involve one or two missed downstream questions after Q1 succeeds. Its largest three-error pattern is \texttt{1000} (Q2-Q4 errors, 5.5\%), where the model detects artifact presence but misses all downstream questions. Gemma-4-31B-it spreads errors across many patterns, with very few complete failures; its largest single-error pattern is \texttt{1101} (Q3 error, 13.7\%), where Q1, Q2, and Q4 are correct but spatial grounding is wrong. In contrast, GPT-5.4, Kimi~K2.5, GLM-4.6V, and Llama-4-Maverick place much more mass on \texttt{0000} (complete failure), with GLM-4.6V especially concentrated there (96.4\%). These diverse patterns show how \textsc{SalArt-VQA} turns a single artifact-present decision into a diagnosis of which part of the Q1-Q4 chain failed.

\begin{figure}[t]
  \centering
  \includegraphics[width=0.95\textwidth]{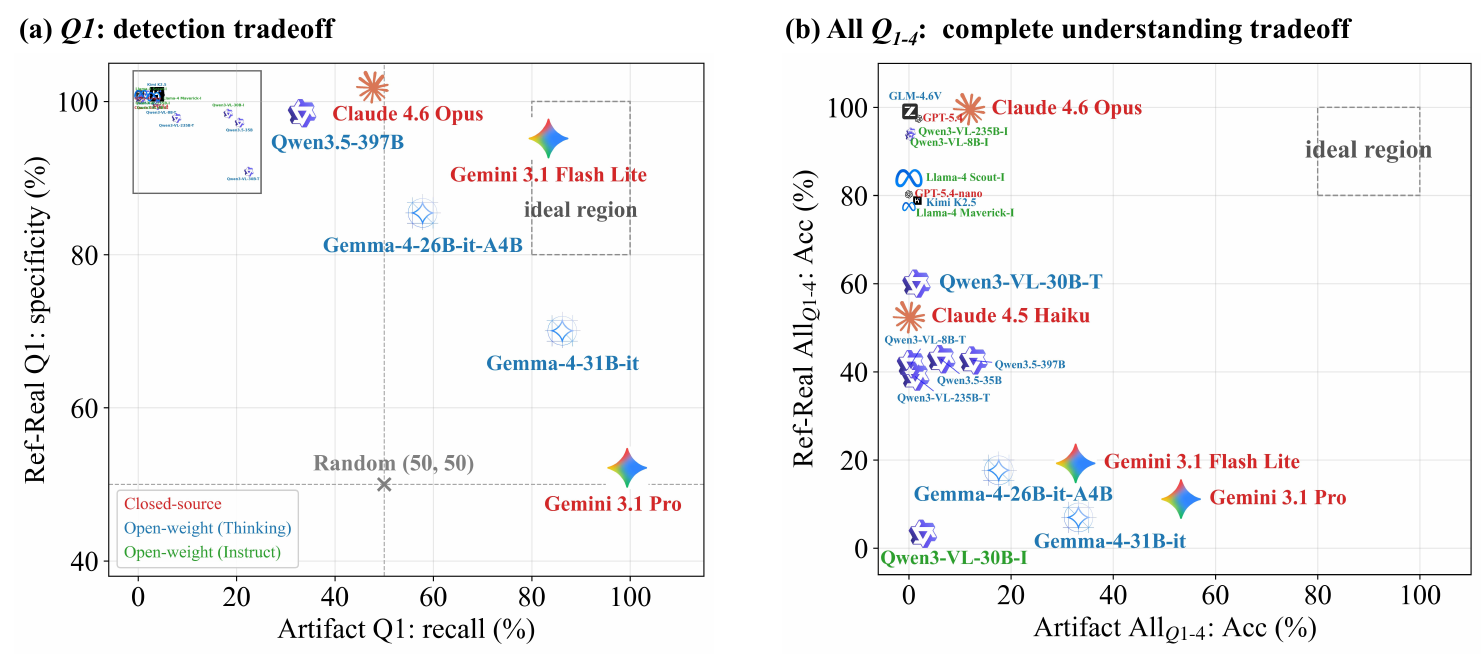}
\caption{\textbf{Artifact detection vs. false-alarm avoidance in \textsc{SalArt-VQA}.} (a) Q1 artifact recall vs. Ref-Real specificity. (b) Artifact All$_{Q1-4}$ accuracy vs. Ref-Real All$_{Q1-4}$ accuracy. 
}
  \label{fig:tradeoff}
\end{figure}

\subsubsection{Behavior Across Artifact and Reference Data}

We next compare the artifact and reference-image blocks of Table~\ref{tab:main}. Human raters are near-perfect across all splits, confirming that the MCQs are answerable, while random guessing is near zero for all-correct metrics. In contrast, \textbf{current VLMs do not jointly achieve artifact-side understanding and reference-side abstention.} The tradeoff becomes clearer when we separate the binary Q1 decision from the full Q1-Q4 diagnostic chain.

\textbf{Q1 detection exposes a sensitivity-specificity tradeoff.} Figure~\ref{fig:tradeoff}(a) shows this binary detection regime. Gemini~3.1~Pro is the most sensitive detector, with 99.4\% artifact Q1 recall, but this comes with only 50.3\% Ref-Real Q1 specificity. GLM-4.6V, Llama-4-Scout-I, Qwen3-VL-235B-A22B-I, and GPT-5.4 occupy the opposite corner: they almost always reject artifacts on Ref-Real images, but rarely detect artifacts when they are present. Gemini~3.1~Flash~Lite is an important intermediate case. It is not the best model on full artifact understanding, but at the binary level it gives the strongest Q1 operating point, with high recall and high specificity. Appendix Table~\ref{tab:q1_det} reports the full precision, recall, specificity, and F1 breakdown behind these groups.

\textbf{The full diagnostic chain reveals a sharper failure mode.} Figure~\ref{fig:tradeoff}(b) shows that the upper-right region is empty once we require all four artifact-side questions and all four Ref-Real questions to be correct. Models that perform best on artifact images still fail to abstain reliably on references. Gemini~3.1~Pro, Gemini~3.1~Flash~Lite, and Gemma-4-31B-it obtain the top artifact All$_{Q1-4}$ scores, but only 7.9\%, 16.0\%, and 3.9\% Ref-Real All$_{Q1-4}$, respectively, and all three score 0.0\% on Ref-PG All$_{Q2-4}$. Thus, even when a model has reasonable Q1 specificity, it may still select a non-E region, box, or defect description on an artifact-free reference.

\textbf{Reference-side calibration can also hide artifact insensitivity.} GLM-4.6V, Qwen3-VL-235B-A22B-I, Qwen3-VL-8B-I, and Llama-4-Scout-I are strongest on artifact-free references, with 68.9-84.9\% Ref-PG All$_{Q2-4}$ and at least 82.3\% Ref-Real All$_{Q1-4}$. However, their artifact All$_{Q1-4}$ scores remain near the 0.4\% random baseline. These models achieve reference-side calibration largely by avoiding artifact commitments.

\textbf{Ref-Real and Ref-PG test different abstention behaviors.} For highly artifact-sensitive models such as Gemini~3.1~Pro and Gemma-4-31B-it, the two reference splits show a similar trend: these models struggle to abstain on both Ref-Real and Ref-PG. For moderately artifact-sensitive models, however, abstention depends more strongly on image type. Claude~Opus~4.6 is the clearest example: it reaches 96.3\% All$_{Q2-4}$ on Ref-Real, yet only 0.8\% All$_{Q2-4}$ on Ref-PG. The Qwen3.5 checkpoints and Gemma-4-26B-it-A4B show the same pattern at lower Ref-Real accuracy. This indicates that rejecting artifacts in ordinary real photographs does not guarantee rejection of the same unsupported artifact hypothesis in a paired generated scene.

\subsection{Additional Analyses and Discussion}
\label{sec:additional-baselines}

\begin{figure}[t]
  \centering
  \includegraphics[width=\textwidth]{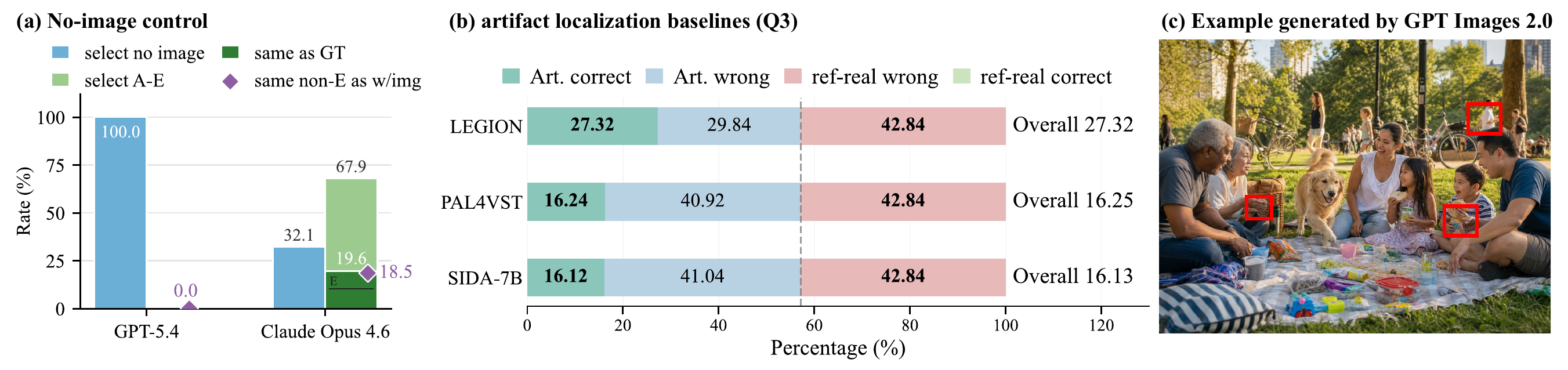}
\caption{\textbf{Additional analyses. }(a) No-image control for Q4: whether models abstain (F) or return A--E without seeing the image. The black line in GT-match bar marks E matches. \textcolor[HTML]{8E5EA2}{$\blacklozenge$} marks the rate at which no-image and with-image choose the same non-E answer. (b) Task-adapted Q3 evaluation of artifact-localization baselines, decomposed into artifact-image correctness/errors and clean-reference correctness/errors. (c) Salient artifact detection remains relevant for modern generators: a ChatGPT Image 2 sample whose overall scene is plausible but still contains clear local irregularities (red boxes).}
  \label{fig:other}
\end{figure}

\textbf{No Image Control.} To check whether Q4 answers depend on the image, we re-run Q4 without the image, keeping A--E and adding a no-image abstention option F (``no image was provided''); the correct answer is always F. Figure~\ref{fig:other}(a) reports all rates over the same 950-image denominator: selection rates for F and A-E, GT matches, and repeats of the with-image answer after excluding E-to-E matches. We evaluate GPT-5.4 and Claude~Opus~4.6. GPT-5.4 selects F on 100.0\% of queries, consistent with its conservative behavior on the main benchmark. Claude~Opus~4.6 abstains on only 32.1\% and returns A--E rather than F on the remaining 67.9\% without any visual input, but these no-image answers match the Q4 ground truth on only 19.6\% of images, with 10.3\% coming from E matches. They also repeat a non-E with-image answer on 18.5\% of images, close to a random same-letter baseline among four choices. Thus, the Q4 options alone do not recover the correct answer. This complements the Ref-PG control: Ref-PG tests whether a defect claim survives in a paired reference, while the no-image control isolates how much of it survives with no image at all.

\textbf{Task-Adapted Artifact Localization Models.} We evaluate PAL4VST~\citep{zhang2023perceptual}, LEGION~\citep{kang2025legion}, and SIDA-7B~\citep{huang2025sida} on Q3 by converting each predicted mask into a box choice via the candidate box with the highest average mask score. This adaptation chooses among A-D but cannot produce the abstention answer E, so clean real references are incorrect by construction; we report artifact-image box accuracy, clean-reference abstention accuracy, and overall accuracy on the artifact images and ref-clean. Figure~\ref{fig:other}(b) shows that these methods are only a partial baseline: LEGION is the best of the three but still selects the correct box on fewer than half of the artifact images, while PAL4VST and SIDA-7B sit close to the forced-choice baseline. All three are inherently miscalibrated on clean references because their mask outputs lack a no-artifact option, clarifying the gap between identifying visible artifacts and knowing when no claim should be made.

\textbf{Is salient artifact detection still meaningful for complex scenes and modern generators?} Although \textsc{SalArt-VQA} deliberately focuses on salient, language-describable artifacts so that each question has a unique visually grounded answer, this controlled scope does not narrow the benchmark's relevance to today's strongest generators. Visible artifacts have not disappeared as image quality has improved: Figure~\ref{fig:other}(c) shows a ChatGPT Image 2~\citep{openai_gpt_image_2} sample where the overall scene is plausible yet several local regions still contain clear irregularities. Complex generated images are themselves built from simpler local components, so reliably detecting, localizing, and abstaining on salient local defects is a prerequisite for analyzing more crowded or ambiguous scenes.

\section{Conclusion}
\label{sec:conclusion}

\textsc{SalArt-VQA} benchmarks the ability of VLMs to analyze salient artifacts and provides a diagnostic evaluation of their artifact claims beyond image-level detection. By decomposing each claim into presence detection, semantic localization, spatial grounding, evidence-grounded defect identification, and reference-side abstention, the benchmark identifies where support for the claim breaks down. Across 20 VLMs, no system reliably completes this chain while remaining calibrated when the queried defect is absent: more sensitive models often make unsupported reference-side claims, while conservative models avoid false alarms largely by missing real artifacts. These results suggest that progress requires defect claims tied more tightly to local visual evidence and evaluation protocols that separate detection, localization, grounding, explanation, and abstention. More broadly, \textsc{SalArt-VQA} can help generative-image pipelines choose artifact evaluators and provide a balanced testbed for VLM sensitivity, grounding, and abstention.

{ 
\small
\bibliographystyle{plainnat}
\bibliography{references}
}


\input{sections/appendix}



\end{document}

%% file: sections/appendix.tex
\appendix

\section{Artifact taxonomy details}
\label{app:taxonomy}

Table~\ref{tab:taxonomy_full} gives the artifact type definitions used by annotators. The categories are organized by the unit being judged: counts of repeated parts, relations between parts, the local quality of one part, the anatomy of a whole body, and scene-level plausibility.

\begin{table}[h]
  \caption{Artifact type definitions provided to annotators.} 
  \label{tab:taxonomy_full}
  \centering
  \scriptsize
    \setlength{\tabcolsep}{1pt}{
  \begin{tabular}{@{}lp{10cm}@{}} 
    \toprule
    \textbf{Type} & \textbf{Definition} \\
    \midrule
    Count anomaly &
    The image has the wrong count for small repeated parts or object components, such as extra or missing fingers, toes, teeth, claws, wheels, handles, or mirrors. Extra or missing major body parts fall under anatomy anomaly. \\
    \addlinespace
    Topology anomaly &
    Parts have an impossible relation: they are merged, disconnected, attached to the wrong place, intersecting, clipping, or violating depth/occlusion order. Use this when connectivity or part-to-part spatial logic is the main defect. \\
    \addlinespace
     Local render/structure anomaly&
    One local part or object is malformed in shape or rendering, while count and attachment are not the main issue. Examples include warped shapes, smeared or under-resolved details, ghost-like endings, blob or stump-like collapse, a fading lower limb, or a hand reduced to an amorphous lump. \\
    \addlinespace
    Anatomy anomaly &
    A biological body has implausible large-scale structure or major-part inventory. This includes extra or missing major parts (arms, legs, heads, tails, wings, necks, trunks, hands, feet, paws, hooves) and impossible whole-body arrangements. \\
    \addlinespace
    Plausibility anomaly &
    The defect concerns scene physics, material behavior, identity, text, or symbols rather than object or body structure. Cases include impossible reflection, shadow, contact, gravity, transparency, or support; a recognizable feature with the wrong identity for its subject; or malformed, unreadable, or semantically wrong text or symbols. Use this only when the other four types do not fit better. \\
    \bottomrule
  \end{tabular}}
\end{table}   

\section{Image generation details}
\label{app:generation} 

\subsection{Generators}

The candidate pool came from three generation sources:

\textbf{Imagen 4.} 
Imagen~4 is an image generation model from Google DeepMind~\citep{googledeepmind2025imagen4}. In our candidate pool, it produced photorealistic images with the fewest salient structural defects.

\textbf{Midjourney.}
Midjourney is a commercial image generation service~\citep{midjourney2026about}. Retained artifacts most often appeared in hands and limbs.

\textbf{FLUX.2 [klein].}
FLUX.2~[klein] is an image generation model from Black Forest Labs~\citep{blackforestlabs2026flux2klein}. We used it across all object classes.

\textbf{Inpainting model.}
For the inpainting path, we used FLUX.1-Fill-dev~\citep{blackforestlabs2026flux1filldev}. The model regenerates a masked region from the text prompt and surrounding pixels.

\textbf{External artifact examples.}
Most retained artifact candidates were generated by the systems above. The 356 Art-DG images also include 30 Midjourney-generated anatomical anomaly examples from HAD~\citep{wang2024detecting}. These images are not a separate split: they are counted inside Art-DG, so the artifact set remains 356 Art-DG images plus 119 Art-PG images. We kept them only when they satisfied the same salient-artifact criteria as the rest of the benchmark, and we authored the \textsc{SalArt-VQA} questions, controls, and answer annotations under the same protocol.

\subsection{Prompt design}
\label{app:generation_prompts}

We adapted prompts from COCO captions~\citep{lin2014microsoft} and wrote them as realistic scene descriptions. We appended ``natural light, detailed surroundings'' and avoided language that asks for defects. We oversampled scene types that often expose structural failures in current generators:

\begin{itemize}
  \item \textbf{Occlusion}: subjects partly hidden by barriers, so the model must complete occluded body parts.
  \item \textbf{Support}: subjects rest on or lean against surfaces, where contact geometry can break.
  \item \textbf{Reflection}: subjects near water, glass, or mirrors, where physical consistency can fail.
  \item \textbf{Crossing}: body parts or objects crossing over each other, often producing topology errors.
  \item \textbf{Contact}: close interaction between subjects or between a subject and an object. 
\end{itemize}

We chose these scene types because prior work reports anatomical inconsistency, reflection errors, compositional relation failures, and implausible scene physics in generated images~\citep{borji2023qualitative,kamali2025characterizing}. The prompts describe plausible real world scenes rather than adversarial inputs.

\subsection{Inpainting pipeline}
\label{app:injection}

The inpainting path is used only for the paired generated split. Its goal is to create one edited image and one unedited counterpart from the same generated scene, so that Q2-Q4 can test the same localized claim on both images.

\textbf{Stage~1: Landmark detection (MediaPipe).}
We use MediaPipe~\citep{lugaresi2019mediapipe} to locate body landmarks. \texttt{FaceLandmarker}, \texttt{HandLandmarker}, and \texttt{PoseLandmarker} provide the keypoints, and we convert them into rough region hints: hand hulls, foot ellipses, and face polygons. Each hint contains a rough boundary, positive prompt points, and optionally negative points or a bounding box.

\textbf{Stage~2: Mask refinement (SAM~2).}
SAM~2~\citep{ravi2024sam} turns each MediaPipe hint into a pixel mask. It uses the positive points, optional negative points, and optional box from Stage~1 to predict candidate masks. We keep masks that cover the seed region and reject masks that spread to unrelated subjects, adjacent objects, or multiple body parts. Small holes are filled before inpainting.

\textbf{Stage~3: Inpainting (FLUX.1-Fill-dev).}
The image and refined mask are passed to FLUX.1-Fill-dev. The prompt is the original scene description and never mentions defects. Each generated reference image is inpainted up to five times with different random seeds. We use guidance scales from 10 to 20, strengths from 0.6 to 0.9, and 20 inference steps as a fixed candidate-generation range: lower strengths often preserve the original part, while higher strengths more often change the local structure without replacing the whole scene.

\textbf{Stage~4: Screening.}
We apply auxiliary VLM screening followed by human verification. The screening script uses InternVL2-2B~\citep{chen2024internvl}, an open-weight VLM outside the evaluated model families, to flag likely artifacts with a short JSON label and reason. This output is only a prefilter for candidate review; final inclusion is decided by human inspection. We keep an inpainted image only if the edited region contains one salient artifact and the rest of the image is visually unchanged. Images that do not yield a qualifying edit after five attempts are excluded. The artifacts come from regenerating fine-grained body structures, such as fingers, toes, and facial features, in a tightly constrained region; the prompts do not ask for errors.

\section{Prompt examples} 
\label{app:prompt_examples}

Table~\ref{tab:prompt_examples} shows one prompt for each subject category. These categories were used to cover common visual subjects during candidate construction; they are not evaluation groups in the main results.

\begin{table}[h]
  \caption{Representative text prompts by subject category.}
  \label{tab:prompt_examples}
  \centering
  \scriptsize
  \begin{tabular}{@{}lp{10.5cm}@{}}
    \toprule
    \textbf{Subject category} & \textbf{Prompt} \\
    \midrule
    Human &
    A realistic photograph of a young commuter stepping into the back seat of a rideshare sedan, one leg already inside, the open door partly blocking the body, natural light, detailed surroundings \\
    \addlinespace
    Animal &
    A realistic photograph of a horse turning through a narrow farm gate, with part of the body hidden by the barrier, natural light, detailed surroundings \\
    \addlinespace
    Objects &
    A realistic photograph of a floor lamp standing on a polished floor near a wall corner, viewed from a low angle, detailed surroundings \\
    \addlinespace
    Sports action &
    A realistic photograph of a gymnast during a balance beam routine, arms extended, natural light, detailed surroundings \\
    \addlinespace
    Indoor/home &
    A realistic photograph of a dining table with place settings for four, one chair slightly pulled out, natural light, detailed surroundings \\
    \addlinespace
    Vehicle &
    A realistic side view photograph of a bicycle parked beside a curb ramp, with one wheel partly hidden by the barrier, no readable text, detailed surroundings \\
    \bottomrule
  \end{tabular}
\end{table}

\section{Annotation process} 
\label{app:annotation}

\subsection{Annotation interface}

Annotators used a Gradio interface that showed the image at full resolution. For each artifact image, they selected the artifact type, wrote the Q2 semantic location, drew the Q3 box, and wrote the Q4 evidence statement. Whenever possible, Q2-Q4 were centered on the same salient artifact; Appendix~\ref{app:hierarchical} uses this alignment for the collapsed diagnostic analysis. For Art-PG images, the paired generated reference image (Ref-PG) was shown beside the edited image to help annotators isolate the changed region.

\subsection{Annotation schema}

The raw annotation payload stores the authored artifact options A-D:

\begin{quote}
\begingroup
\scriptsize
\begin{verbatim}
{
  "artifact_type": "count_anomaly",
  "q1_ref": "yes",
  "semantic_loc": {"ref": "B", "options": {"A": "...", "B": "..."}},
  "box": {"ref": "A", "candidates": {"A": [322, 181, 414, 271]}},
  "evidence": {"ref": "B", "options": {"A": "...", "B": "..."}}
}
\end{verbatim} 
\endgroup
\end{quote}

For Q2-Q4, annotators author options A-D and choose the reference answer. The export step adds the fixed ``none of these'' answer. For Q2 and Q4, the final table includes \texttt{q2\_option\_e} (``none of these regions'') and \texttt{q4\_option\_e} (``none of these descriptions''). For Q3, option~E means ``none of these boxes'' and has no box coordinates. The benchmark therefore scores Q2-Q4 over A-E.

Distractors A-D are plausible by design. They usually put the right defect at the wrong location, the wrong defect at the right location, or a suspicious but non-defective region into the answer set.

\subsection{Location conventions}

Spatial descriptions follow a two-level convention. Subjects are identified from the \emph{viewer's} perspective: ``the person on the left'' is the person on the left side of the image as displayed. Body parts use the subject's own \emph{anatomical} left and right: ``the right hand of the person on the left'' refers to the anatomical right hand of the person appearing on the viewer's left. This convention was established at the start of annotation and enforced through training and interface reminders.

\subsection{Quality control}
Quality control combines consensus filtering and export-time validation. Annotators first received guidelines for the salient-artifact criterion, left/right conventions, option alignment, and box drawing, with pilot examples. Each candidate item was then checked by three reviewers for artifact salience, location clarity, Q2--Q4 alignment, and box precision. We retained only items for which all three reviewers agreed that the annotation was valid; items with disagreement were revised and rechecked or excluded from the release. This consensus protocol ensures that every released item passed unanimous review, although we do not use it to estimate a formal inter-annotator agreement statistic. During export, automated checks reject duplicate row IDs, invalid pair structure, empty reference answers, invalid option labels, non-positive boxes, and boxes outside image bounds.


\section{VQA question design}
\label{app:vqa_design}

\subsection{Question prompts and options}

\begin{figure}[h]
  \centering
  \begin{tabular}{@{}cc@{}}
    \includegraphics[width=0.31\linewidth]{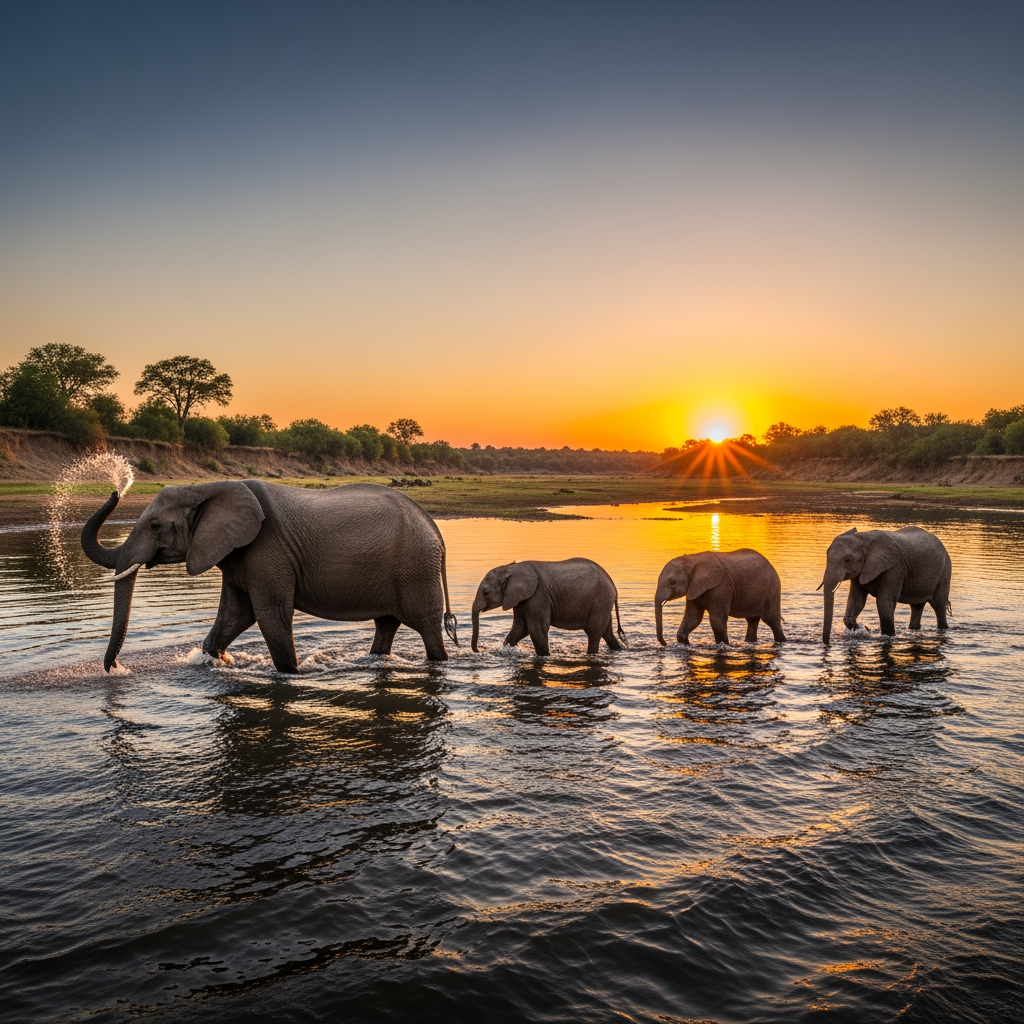} &
    \includegraphics[width=0.31\linewidth]{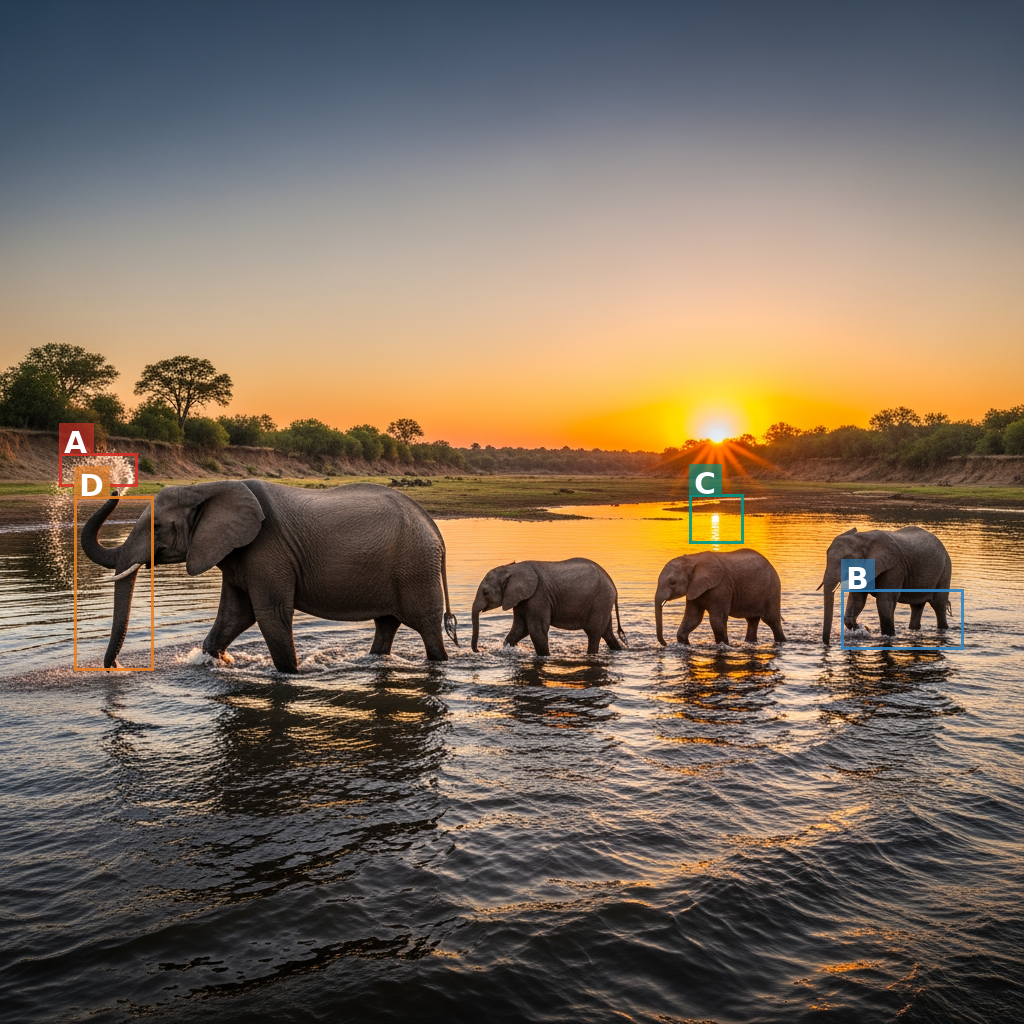} \\
    \scriptsize Original image for Q1, Q2, and Q4. &
    \scriptsize Box overlay image for Q3.
  \end{tabular}
  \caption{Visual inputs used by the four VQA prompts. Q1, Q2, and Q4 are answered from the original image. Q3 is answered from an overlay image containing boxes A-D.}
  \label{fig:app_vqa_prompt_example}
\end{figure}

The benchmark uses four independent, single-turn question prompts. The wording of Q1-Q4 is fixed across the dataset, and Q1 always uses the same yes/no answer set. 
For Q2-Q4, annotators provide the image-specific choices: semantic regions for Q2, boxes for Q3, and evidence statements for Q4. 
Figure~\ref{fig:app_vqa_prompt_example} shows one benchmark instance. Q1, Q2, and Q4 use the original image; Q3 uses the same image with candidate boxes overlaid. All models receive the same pre-rendered Q3 overlay for a given item. Boxes A-D use fixed colors (red, blue, teal, and orange), with line width and bold letter badges scaled by image size; overlapping boxes are left as drawn and therefore identical across models. The prompt text below is shown as the model sees it, with textual choices inline for Q1, Q2, and Q4. For Q3, choices A-D are the labeled boxes in the overlay image, and E is the none-of-these choice described by the prompt. The location convention is intentionally repeated in Q2 and Q4 because each prompt is evaluated independently; we reproduce both copies verbatim below.

\newenvironment{vqaprompt}{%
  \par\smallskip\noindent
  \begingroup
  \fontsize{7}{8.1}\selectfont\sffamily\color{black!72}%
  \setlength{\parindent}{0pt}%
  \setlength{\parskip}{0.34em}%
}{%
  \par\endgroup\vspace{0.25em}
}
\newcommand{\vqaheading}[1]{\par\smallskip\noindent\textbf{#1}\par\vspace{-0.25em}}

\vqaheading{Q1: PresenceQA.}
\begin{vqaprompt}
This image may be a real photograph or an AI-generated image.

If you judge it to be a real photograph, treat it as having no generation artifacts. If you judge it to be an AI-generated image, it may or may not contain visible generation artifacts.

Do not answer yes merely because the image looks AI-generated, stylized, cinematic, or slightly imperfect. 
Conversely, do not answer no merely because a rare real case could have an unusual structure, such as a person naturally having four or fewer fingers.

Does this image contain at least one salient artifact, meaning at least one obvious abnormality that stands out more than minor, weak, or debatable imperfections?

Answer yes only if there is at least one salient artifact. If the image appears real, or if it appears AI-generated but no salient artifact is present, answer no.

Answer with exactly one word: yes or no. 
\end{vqaprompt}

\vqaheading{Q2: SemanticLocQA.}
\begin{vqaprompt}
This image may be a real photograph or an AI-generated image.

If you judge it to be a real photograph, treat it as having no generation artifacts. If you judge it to be an AI-generated image, it may or may not contain visible generation artifacts.

Among the provided options, which object, body part, local region, or larger subject region best captures the most salient artifact?

Judge only among the provided options.

If you judge the image to be real, choose E. If you judge the image to be AI-generated but none of options A-D correctly identifies a region that actually contains a visible artifact, choose E.

Choose A-D only if you judge that the selected option actually contains a visible artifact. Do not reject a region solely because an unusual real-world case could make the appearance possible, such as a person naturally having four or fewer fingers. If more than one option could contain a visible artifact, choose the one that best captures the most visually and semantically salient artifact rather than a smaller, weaker, or secondary issue.

Resolve location wording in two steps:

1. Use viewer perspective only to identify which person, animal, or object is being referred to.

2. After that subject is identified, use that subject's own anatomical left/right for body parts.

Important:

- Do not determine left/right from where the body part appears on the screen.

- A hand that appears on the viewer's left side of the image can still be the subject's right hand, and a hand that appears on the viewer's right side of the image can still be the subject's left hand.

- First decide which subject is being referred to, then decide that subject's own anatomical left/right.

- Do not flip left/right to match the viewer's perspective, and do not relabel the option text.

For subject identification, phrases such as 'leftmost person', 'rightmost person', 'first person from the right', 'second person from the left', and 'second person from the right' are all based on the viewer's perspective. If there are three people, 'the middle person' or 'the person in the center' means the person between the leftmost and rightmost people.

After the subject is fixed, phrases such as 'left hand', 'right hand', 'left leg', 'right leg', 'left eye', and 'right eye' are based on that subject's own body, not the viewer's mirrored view.

Examples: 'the left hand of the rightmost person', 'the right hand of the second person from the left', 'the left leg of the middle person', and 'the right eye of the person in the center' all use the subject's own anatomy after the subject is identified.

Example: 'the right hand of the person on the right' means first find the person on the right from the viewer's perspective, then inspect that person's own right hand.

Example: 'the left hand of the person on the right' does not mean the hand appearing on the viewer's left side of that person; it means that subject's own left hand.

Example: 'the right hand of the leftmost person' may visually appear on either side of that person in the image, depending on pose and camera angle.

Example: 'the left leg of the middle person' means the middle person's own left leg, not whichever leg appears on the viewer's left.

Do not rewrite the option text, do not flip left/right to match the image, and choose only from the provided options.

Options:

A. the water spray from the leading elephant's trunk

B. the legs of the elephant on the far right

C. the reflection of the sun on the water

D. The trunk of the elephant on the far left

E. none of these regions

Answer with exactly one uppercase letter:

A or B or C or D or E
\end{vqaprompt}

\vqaheading{Q3: BoxGroundingQA.}
\begin{vqaprompt}
This image may be a real photograph or an AI-generated image.

If you judge it to be a real photograph, treat it as having no generation artifacts. If you judge it to be an AI-generated image, it may or may not contain visible generation artifacts.

Among the provided boxes, which box best captures the most salient artifact?

Judge only among the provided boxes.

If you judge the image to be real, choose E. If you judge the image to be AI-generated but none of boxes A-D correctly captures a visible artifact-bearing region, choose E.

Choose A-D only if you judge that the selected box actually contains a visible artifact. Do not reject a box solely because the boxed appearance might have a rare real-world explanation, such as a person naturally having four or fewer fingers. If more than one box could contain a visible artifact, choose the one that best captures the most visually and semantically salient artifact rather than a smaller, weaker, or secondary issue.

Answer with exactly one uppercase letter:

A or B or C or D or E
\end{vqaprompt}

\vqaheading{Q4: EvidenceQA.}
\begin{vqaprompt}
This image may be a real photograph or an AI-generated image.

If you judge it to be a real photograph, treat it as having no generation artifacts. If you judge it to be an AI-generated image, it may or may not contain visible generation artifacts.

Which option best describes the most salient artifact?

Judge only among the provided options.

If you judge the image to be real, choose E. If you judge it to be AI-generated but none of options A-D correctly describes a visible artifact, choose E.

Choose A-D only if the selected description is directly supported by visible evidence and best matches the most salient artifact in the image. Do not reject a description solely because a rare real case could have a similar structure, such as a person naturally having four or fewer fingers. If more than one option describes a visible artifact, choose the one that best matches the salient artifact rather than a smaller, weaker, or secondary issue.

Do not choose an option that is merely true about the scene, merely suggests that the image looks AI-generated, or describes a weak or debatable irregularity that does not rise to the level of the salient artifact.

If an evidence option uses location wording, apply the same location convention below.

Resolve location wording in two steps:

1. Use viewer perspective only to identify which person, animal, or object is being referred to.

2. After that subject is identified, use that subject's own anatomical left/right for body parts.

Important:

- Do not determine left/right from where the body part appears on the screen.

- A hand that appears on the viewer's left side of the image can still be the subject's right hand, and a hand that appears on the viewer's right side of the image can still be the subject's left hand.

- First decide which subject is being referred to, then decide that subject's own anatomical left/right.

- Do not flip left/right to match the viewer's perspective, and do not relabel the option text.

For subject identification, phrases such as 'leftmost person', 'rightmost person', 'first person from the right', 'second person from the left', and 'second person from the right' are all based on the viewer's perspective. If there are three people, 'the middle person' or 'the person in the center' means the person between the leftmost and rightmost people.

After the subject is fixed, phrases such as 'left hand', 'right hand', 'left leg', 'right leg', 'left eye', and 'right eye' are based on that subject's own body, not the viewer's mirrored view.

Examples: 'the left hand of the rightmost person', 'the right hand of the second person from the left', 'the left leg of the middle person', and 'the right eye of the person in the center' all use the subject's own anatomy after the subject is identified.

Example: 'the right hand of the person on the right' means first find the person on the right from the viewer's perspective, then inspect that person's own right hand.

Example: 'the left hand of the person on the right' does not mean the hand appearing on the viewer's left side of that person; it means that subject's own left hand.

Example: 'the right hand of the leftmost person' may visually appear on either side of that person in the image, depending on pose and camera angle.

Example: 'the left leg of the middle person' means the middle person's own left leg, not whichever leg appears on the viewer's left.

Do not rewrite the option text, do not flip left/right to match the image, and choose only from the provided options.

Options:

A. the water spray appears as a solid, unnatural shape rather than droplets

B. the legs of the elephant on the far right appear to float without proper water displacement

C. the reflection of the sun on the water is perfectly sharp and static

D. the elephant on the far left has an extra trunk

E. none of these descriptions

Answer with exactly one uppercase letter:

A or B or C or D or E
\end{vqaprompt}

\subsection{Distractor design}

Q2-Q4 options are written around the same candidate regions. Q2 names a region, Q3 marks the corresponding box, and Q4 pairs that region with a defect claim. With this construction, a Q4 success followed by a Q2 or Q3 error is a localization failure, not a mismatch in option wording. For Q2, distractors vary the referred subject and local part, such as the left person versus the right person, or a hand versus a sleeve. The model must identify both the subject and the affected part. For Q3, distractor boxes cover plausible but wrong regions: the right part on the wrong subject, a nearby unaffected part, or a visually suspicious clean region. All boxes have positive area and lie within image bounds. For Q4, distractors combine the wrong location with the right defect type, the right location with the wrong defect type, or a plausible defect not supported by the image.

\textbf{Cross-question consistency.} The collapsed diagnosis in Appendix~\ref{app:hierarchical} uses this per-option consistency. The correct-answer letter is identical across Q2-Q4 for 97.3\% of artifact images. In the remaining 2.7\%, the options still refer to matched candidate regions, but annotation review selected a different letter for the best semantic phrase, box, or evidence statement. Scoring always uses the reference answer for each individual question; the diagnostic table uses those per-question correctness values rather than forcing a shared letter.

\subsection{Scoring}

All questions use exact-match scoring. Q1 accepts ``yes'' or ``no'' case-insensitively. Q2-Q4 accept one letter, A-E. We give no partial credit, so scoring does not depend on fuzzy matching or manual judgment of free-form text.

\subsection{Evaluation protocol}

Models are evaluated in a single-turn, zero-shot setting. Each question is sent in an independent API call with the fixed prompt wording above and the options for that image. The image is provided with the question text. For Q3, the visual input also includes an overlay with boxes A-D.

\textbf{Question assignment by image category.}
The 475 artifact images and 356 Ref-Real images receive Q1-Q4. The Ref-Real images have answer Q1\,=\,``no'' and Q2-Q4\,=\,E. The 119 Ref-PG images from the pixel-aligned inpainting pairs receive only Q2-Q4, with answer~E for each question. We skip Q1 on this split because Ref-PG images are still generated images: minor unrelated synthesis flaws can blur a global artifact-present judgment. Q2-Q4 are tied to the paired salient defect. A non-E answer on Ref-PG asserts that the artifact-side region, box, or defect description is still supported after the edit is removed, so minor unrelated flaws elsewhere do not make an A-D answer correct.

Each pixel-aligned pair contains one Art-PG image and one Ref-PG image. Both use the same Q2-Q4 options A-D. In Art-PG, one option points to the edited defect. In Ref-PG, that defect is absent and the other A-D options are still distractors. Thus all A-D claims about the paired salient artifact are false on Ref-PG, and the correct answer is~E. This is the paired hallucination control used in Section~\ref{sec:vqa}: a non-E answer on Ref-PG is a false artifact claim under the same options.

\section{Additional results}
\label{app:additional}

\subsection{Hierarchical analysis}
\label{app:hierarchical}

Table~\ref{tab:diagnostic} gives an appendix-only hierarchical breakdown for the 475 artifact images. This four-bucket view collapses the 16 Q1-Q4 answer patterns in Figure~\ref{fig:artifact_diagnostic}(b). We first separate Q1 misses. Among the remaining images, we next check Q4 because it is the most semantically complete downstream question: the model must select the defect description supported by the image. If Q1 and Q4 are correct but Q2 or Q3 is wrong, we count the image as a localization failure. This ordering is a diagnostic summary of the answer pattern, not an additional task protocol.

\begin{table}[t]
  \caption{Hierarchical failure diagnosis on representative models over 475 artifact images (\%). Rows span the strongest artifact-side models and the high-specificity, low-recall operating points discussed in the main results. Each image is assigned to exactly one category. Q1 miss means the artifact-side Q1 answer is incorrect. Evidence failure means Q1 is correct but Q4 is wrong. Localization failure means Q1 and Q4 are correct but Q2 or Q3 (or both) is wrong. }
  \label{tab:diagnostic}
  \centering
  \small
  \begin{tabular}{@{}l rrrr@{}}
    \toprule
    Model & \makecell{Q1\\miss} & \makecell{Evidence\\failure} & \makecell{Localization\\failure} & \makecell{Full\\success} \\
    \midrule
    Gemini 3.1 Pro & 0.6 & 18.9 & 27.2 & 53.3 \\
    Gemini 3.1 Flash Lite & 16.6 & 32.4 & 18.3 & 32.6 \\
    Gemma-4-31B-it (thinking) & 13.7 & 25.9 & 27.2 & 33.3 \\
    Claude 4.6 Opus & 52.4 & 23.8 & 12.0 & 11.8 \\
    Qwen3.5-397B-A17B & 66.7 & 10.5 & 10.1 & 12.6 \\
    GPT-5.4 & 95.4 & 1.3 & 1.5 & 1.9 \\
    GLM-4.6V (thinking) & 99.6 & 0.2 & 0.0 & 0.2 \\
    Qwen3-VL-235B-A22B-I & 98.5 & 0.4 & 0.6 & 0.4 \\
    Qwen3-VL-30B-A3B-I & 81.7 & 10.9 & 4.6 & 2.7 \\
    \bottomrule
  \end{tabular}
\end{table}

For the three strongest artifact-side models, Q1 is no longer the main bottleneck. Gemini~Pro misses only 0.6\% of artifact images at Q1, then loses 18.9\% to evidence failures and 27.2\% to localization failures. Flash~Lite and Gemma-4-31B-it detect less often, but their remaining errors also occur mainly after the model has already said that an artifact is present. Claude~Opus~4.6 and Qwen3.5-397B-A17B have a different shape: Q1 is still the main filter, and the model only sometimes reaches the downstream questions. High-specificity, low-recall models are mostly abstainers on artifact images. GPT-5.4, GLM-4.6V, Kimi~K2.5, Llama-4 Scout/Maverick, and the conservative Qwen3-VL instruct checkpoints get strong reference-side scores largely by avoiding artifact commitments. Qwen3-VL-30B-A3B-I is the edge case: it reaches 18.3\% Q1 recall, but only 2.7\% artifact all-4 and 0.0\% Ref-Real all-4.

This table should not be read as a causal chain. Since Q1-Q4 are asked independently, a Q1 miss does not always mean the model has no local visual signal. Qwen3.5-35B-A3B has 20.6\% Q1 recall but 51.6\% Q4 accuracy; 37.5 points of that Q4 accuracy come from images where Q1 was wrong. Qwen3.5-397B-A17B and GPT-5.4 show the same pattern. Q1 is therefore a commitment threshold as well as a perception test. The downstream questions separate skills that aggregate accuracy would merge. Table~\ref{tab:type_contingency} shows Q4-only and Q3-only disagreements on the same images, plus Q2/Q3 disagreements between semantic and box localization. The type block gives a separate check on artifact categories: Gemini, Gemma, and Qwen3.5-397B are higher on plausibility than local render/structure, while Claude~Opus~4.6 reverses that pattern. The category effect is therefore model-specific, not a universal difficulty ranking.

Finally, the unsolved images are not isolated failures of one model. A per-image oracle over all 20 models solves 327/475 artifact images, or 68.8\%. The remaining 148 images are not solved by any evaluated model under all-4 correctness, giving a rough ceiling for the current evaluated model pool rather than a benchmark upper bound.

\subsection{Q1 detection metrics}
\label{app:q1_det}

Table~\ref{tab:q1_det} reports the Q1 detection metrics derived from artifact images as positives and Ref-Real images as negatives. These numbers complement the Q1 accuracies in Table~\ref{tab:main} by making the precision-recall-specificity tradeoff explicit.

\begin{table}[h]
  \caption{Q1 detection metrics on artifact versus Ref-Real images (\%). Precision, recall, and specificity are computed from Q1 with artifact images as positives and Ref-Real images as negatives. Row groups summarize observed operating points for readability and are not additional model categories.}
  \label{tab:q1_det}
  \centering
  \scriptsize
  \begin{tabular}{@{}l rrrr@{}}
    \toprule
    Model & Precision & Recall & Specificity & F1 \\
    \midrule
    \multicolumn{5}{@{}l}{\textbf{High artifact recall / all-4}} \\
    Gemini 3.1 Pro & 72.7 & 99.4 & 50.3 & 84.0 \\
    Gemini 3.1 Flash Lite & 94.3 & 83.4 & 93.3 & 88.5 \\
    Gemma-4-31B-it & 78.4 & 86.3 & 68.3 & 82.2 \\
    \addlinespace
    \multicolumn{5}{@{}l}{\textbf{Moderate artifact recall / all-4}} \\
    Claude 4.6 Opus & 100.0 & 47.6 & 100.0 & 64.5 \\
    Gemma-4-26B-it-A4B & 82.6 & 57.9 & 83.7 & 68.1 \\
    Qwen3.5-397B-A17B & 92.9 & 33.3 & 96.6 & 49.0 \\
    Qwen3.5-35B-A3B & 88.3 & 20.6 & 96.3 & 33.4 \\
    \addlinespace
    \multicolumn{5}{@{}l}{\textbf{High specificity, low artifact recall}} \\
    GPT-5.4 & 100.0 & 4.6 & 100.0 & 8.9 \\
    GLM-4.6V & 100.0 & 0.4 & 100.0 & 0.8 \\
    Qwen3-VL-235B-A22B-I & 100.0 & 1.5 & 100.0 & 2.9 \\
    \makecell[l]{Llama-4-Scout-17B-16E-I} & 100.0 & 0.6 & 100.0 & 1.3 \\
    Qwen3-VL-8B-I & 50.0 & 0.4 & 99.4 & 0.8 \\
    \makecell[l]{Llama-4-Maverick-17B-128E-I} & 93.8 & 3.2 & 99.7 & 6.1 \\
    Kimi K2.5 & 100.0 & 3.8 & 100.0 & 7.3 \\
    GPT-5.4-nano & 50.0 & 0.2 & 99.7 & 0.4 \\
    \addlinespace
    \multicolumn{5}{@{}l}{\textbf{Low end-to-end success}} \\
    Qwen3-VL-30B-A3B-I & 90.6 & 18.3 & 97.5 & 30.5 \\
    Qwen3-VL-30B-A3B-T & 74.8 & 22.5 & 89.9 & 34.6 \\
    Qwen3-VL-235B-A22B-T & 77.1 & 7.8 & 96.9 & 14.1 \\
    Qwen3-VL-8B-T & 78.3 & 3.8 & 98.6 & 7.2 \\
    Claude 4.5 Haiku & 50.0 & 0.4 & 99.4 & 0.8 \\
    \bottomrule
  \end{tabular}
\end{table}

\subsection{Artifact performance by construction route}
\label{app:path_split}

Table~\ref{tab:path_split} separates artifact performance on the 356 Art-DG images and the 119 Art-PG images. The two subsets differ in how artifacts are produced: Art-DG comes from direct generation and filtering, while Art-PG comes from localized inpainting failures on paired generated reference images.
\begin{table}[h]
  \caption{Artifact performance split by artifact-image source (\%). Art-DG contains the 356 directly curated artifact images; Art-PG contains the 119 inpainting-induced artifact images. We report Q1 detection accuracy and artifact all-4 for each subset. Row groups summarize observed operating points for readability and are not additional model categories.}
  \label{tab:path_split}
  \centering
  \scriptsize
  \begin{tabular}{@{}l rrrr@{}}
    \toprule
    Model & Art-DG Q1 & Art-PG Q1 & Art-DG All-4 & Art-PG All-4 \\
    \midrule
    \multicolumn{5}{@{}l}{\textbf{High artifact recall / all-4}} \\
    Gemini 3.1 Pro & 99.4 & 99.2 & 53.1 & 53.8 \\
    Gemini 3.1 Flash Lite & 78.9 & 96.6 & 30.1 & 40.3 \\
    Gemma-4-31B-it & 87.6 & 82.4 & 35.4 & 26.9 \\
    \addlinespace
    \multicolumn{5}{@{}l}{\textbf{Moderate artifact recall / all-4}} \\
    Claude 4.6 Opus & 44.7 & 56.3 & 9.6 & 18.5 \\
    Gemma-4-26B-it-A4B & 58.7 & 55.5 & 18.0 & 16.8 \\
    Qwen3.5-397B-A17B & 36.0 & 25.2 & 14.9 & 5.9 \\
    Qwen3.5-35B-A3B & 22.5 & 15.1 & 7.0 & 4.2 \\
    \addlinespace
    \multicolumn{5}{@{}l}{\textbf{High specificity, low artifact recall}} \\
    GPT-5.4 & 5.9 & 0.8 & 2.5 & 0.0 \\
    GLM-4.6V & 0.6 & 0.0 & 0.3 & 0.0 \\
    Qwen3-VL-235B-A22B-I & 1.1 & 2.5 & 0.3 & 0.8 \\
    \makecell[l]{Llama-4-Scout-17B-16E-I} & 0.6 & 0.8 & 0.0 & 0.0 \\
    Qwen3-VL-8B-I & 0.3 & 0.8 & 0.3 & 0.0 \\
    \makecell[l]{Llama-4-Maverick-17B-128E-I} & 3.9 & 0.8 & 0.0 & 0.0 \\
    Kimi K2.5 & 3.9 & 3.4 & 2.2 & 0.0 \\
    GPT-5.4-nano & 0.3 & 0.0 & 0.0 & 0.0 \\
    \addlinespace
    \multicolumn{5}{@{}l}{\textbf{Low end-to-end success}} \\
    Qwen3-VL-30B-A3B-I & 17.1 & 21.8 & 3.4 & 0.8 \\
    Qwen3-VL-30B-A3B-T & 17.7 & 37.0 & 1.7 & 0.8 \\
    Qwen3-VL-235B-A22B-T & 7.6 & 8.4 & 1.7 & 0.0 \\
    Qwen3-VL-8B-T & 3.7 & 4.2 & 0.6 & 0.0 \\
    Claude 4.5 Haiku & 0.6 & 0.0 & 0.0 & 0.0 \\
    \bottomrule
  \end{tabular}
\end{table}
The two construction routes do not produce a single difficulty ordering. Gemini~Pro is nearly unchanged across Art-DG and Art-PG, with all-4 accuracy around 53\% on both. Flash~Lite and Claude~Opus~4.6 do better on Art-PG, where the salient defect is concentrated in one edited region, while Gemma-4-31B-it and Qwen3.5-397B-A17B do better on Art-DG. The combined artifact block in Table~\ref{tab:main} represents an average of two complementary routes, rather than a result from a single source.



\subsection{Artifact type and cross-question checks}

Table~\ref{tab:type_contingency} keeps the appendix diagnostics that are not visible from aggregate accuracy alone. The type block reports all-4 accuracy within the displayed artifact categories. The disagreement block shows asymmetric errors between aligned questions: Q4-only means the evidence statement is correct but the Q3 box is wrong; Q3-only means the reverse. The Q2/Q3 columns make the same comparison for semantic and box localization.

\begin{table*}[t]
  \caption{Artifact-type all-4 accuracy and cross-question disagreement rates on 475 artifact images (\%). Type columns report all-4 accuracy within each displayed artifact type. Disagreement columns report asymmetric per-image correctness patterns over all artifact images. Rows report representative nontrivial artifact-side operating points from Table~\ref{tab:diagnostic}, plus GPT-5.4 as a high-specificity, low-recall reference point; near-zero rows are omitted for readability.}
  \label{tab:type_contingency}
  \centering
  \scriptsize
  \begin{tabular}{@{}l rrrr rr rr@{}}
    \toprule
    & \multicolumn{4}{c}{Artifact type all-4} & \multicolumn{2}{c}{Q4 vs. Q3} & \multicolumn{2}{c}{Q2 vs. Q3} \\
    \cmidrule(lr){2-5} \cmidrule(lr){6-7} \cmidrule(lr){8-9}
    Model & Anatomy & Topology & \makecell{Local\\render} & Plaus. & Q4-only & Q3-only & Q2-only & Q3-only \\
    \midrule
    Gemini 3.1 Pro & 51.0 & 62.4 & 42.4 & 64.1 & 18.7 & 9.5 & 15.4 & 10.7 \\
    Gemini 3.1 Flash Lite & 20.3 & 45.6 & 22.0 & 38.5 & 14.7 & 20.4 & 13.5 & 13.5 \\
    Gemma-4-31B-it & 35.0 & 34.4 & 25.4 & 51.3 & 28.0 & 13.5 & 24.0 & 9.7 \\
    Claude 4.6 Opus & 7.7 & 13.6 & 20.3 & 10.3 & 19.2 & 23.4 & 22.9 & 10.1 \\
    Qwen3.5-397B-A17B & 12.6 & 13.6 & 5.1 & 41.0 & 23.2 & 17.5 & 19.6 & 10.7 \\
    GPT-5.4 & 2.1 & 2.4 & 0.0 & 7.7 & 7.6 & 25.3 & 12.4 & 15.6 \\
    \bottomrule
  \end{tabular}
\end{table*}  

\section{Broader impacts and ethics statement} 
\label{app:impacts}

\textsc{SalArt-VQA} is an evaluation benchmark for diagnosing whether VLMs can detect, localize, and reject salient artifact claims in generated images. The intended use is model auditing. Better artifact evaluation can help developers find cases where a model over-claims visual defects, misses obvious synthesis errors, or fails to abstain on clean references. The benchmark images are synthetic or generated controls rather than photographs collected from private users. Some images contain people or person-like subjects, but the annotations describe visible artifact regions and defect evidence rather than personal identity or protected attributes. The benchmark is not designed for identity recognition, demographic inference, or surveillance. Annotators were instructed to judge visual support for artifact claims, not to label demographic categories. The paired reference setting directly tests unsupported artifact claims: on Ref-PG, all A-D options are false with respect to the paired salient defect, so a correct model must choose the abstention option. We frame the benchmark as an evaluation tool, report aggregate model behavior, and use closed-set questions rather than open-ended instructions for producing artifacts.

\section{Reproducibility statement}
\label{app:Reproducibility}

The paper gives the full VQA prompt wording, answer schema, scoring rule, and question assignment protocol in Appendix~\ref{app:vqa_design}. Appendix~\ref{app:generation} and Appendix~\ref{app:injection} describe the two artifact construction routes, including the generator families, masking pipeline, inpainting model, sampling ranges, and screening procedure. The main text reports the model list, zero-shot protocol, answer parsing rule, and evaluation window for closed-source systems.

For evaluation and data construction, we use official provider APIs for closed-source systems and Hugging Face implementations for open-weight models when available. For closed-source models, the evaluation window is 2026-04-08 to 2026-04-10 (UTC).

\section{Limitations}
\label{app:lim}

\textsc{SalArt-VQA} is a controlled diagnostic benchmark rather than a survey of how often artifacts appear in the wild. Each artifact image is built around one salient, language-describable defect so that Q1-Q4 have well-defined answers. This choice makes the benchmark useful for separating detection, semantic localization, box grounding, evidence selection, and abstention. It also means that the benchmark should not be read as covering every artifact that can occur in generated images. At the same time, many artifacts in complex generated scenes are built from local, visually identifiable defects of the kind studied here. Therefore, reliably detecting, localizing, and abstaining on these controlled salient artifacts is an important prerequisite for analyzing more crowded, ambiguous, or open-ended generated images.


\section{Compute resources}
\label{app:resources}

Local experiments ran on a workstation with two NVIDIA RTX A6000 GPUs. We ran FLUX.2 generation, FLUX.1-Fill-dev inpainting, SAM~2 mask refinement, InternVL2-2B filtering, and smaller open-weight VLM evaluations locally. Examples of locally evaluated VLMs include Gemma-4-31B-it and Qwen3.5-35B-A3B. Closed-source models and larger open-weight models that did not fit in local memory were queried through official APIs. We did not train or fine-tune VLMs for this paper.